\documentclass[11pt]{article}

\usepackage[final]{acl}

\usepackage{times}
\usepackage{latexsym}

\usepackage[T1]{fontenc}

\usepackage[utf8]{inputenc}

\usepackage{microtype}

\usepackage{inconsolata}

\usepackage{graphicx}

\usepackage{booktabs} 
\usepackage{multirow}
\usepackage{subcaption}  
\usepackage{amssymb}
\usepackage[normalem]{ulem}
\useunder{\uline}{\ul}{}
\usepackage{ulem}
\usepackage{textcomp} 
\usepackage{arydshln} 
\usepackage{pifont} 

\usepackage[table, svgnames]{xcolor}
\definecolor{MyTurquoise}{HTML}{05D6A0}
\usepackage{pgfplots}
\pgfplotsset{compat=1.18}
\usepackage{tikz}
\usepackage{array}
\usepackage{marvosym}

\newcommand{\accHeat}[1]{%
\begingroup
\edef\temp{#1}%
\def\dash{-}%
\ifx\temp\dash
    \cellcolor{green!45}{-}%
\else
    \pgfmathsetmacro{\v}{#1}%
    \pgfmathparse{\v==100 ? 1 : 0}%
    \ifnum\pgfmathresult>0
      \cellcolor{green!45}{#1}%
\else
    \pgfmathparse{\v>90 ? 1 : 0}%
    \ifnum\pgfmathresult>0
    \pgfmathsetmacro{\p}{(\v-60)/(100 - 60)*40}%
    \pgfmathtruncatemacro{\pp}{max(0,min(100,\p))}%
    \edef\tempcolor{green!\pp}%
    \expandafter\cellcolor\expandafter{\tempcolor}{#1}%
\else
    \pgfmathparse{\v>60 ? 1 : 0}%
    \ifnum\pgfmathresult>0
    \pgfmathsetmacro{\p}{(\v-60)/(100 - 60)*40}%
    \pgfmathtruncatemacro{\pp}{max(0,min(100,\p))}%
    \edef\tempcolor{green!\pp!yellow}%
    \expandafter\cellcolor\expandafter{\tempcolor}{#1}%
\else
    \pgfmathparse{\v>40 ? 1 : 0}%
    \ifnum\pgfmathresult>0
    \pgfmathsetmacro{\p}{(\v-40)/(100-40)*40}%
    \pgfmathtruncatemacro{\pp}{max(0,min(100,\p))}%
    \edef\tempcolor{yellow!\pp!orange}%
    \expandafter\cellcolor\expandafter{\tempcolor}{#1}%
\else
    \pgfmathsetmacro{\p}{\v/60*50}%
    \pgfmathtruncatemacro{\pp}{max(0,min(100,\p))}%
    \edef\tempcolor{orange!\pp!red!60}%
    \expandafter\cellcolor\expandafter{\tempcolor}{#1}%
\fi
\fi
\fi
\fi
\fi
\endgroup
}

\title{CNSL-bench: Benchmarking the Sign Language Understanding Capabilities of MLLMs on Chinese National Sign Language}

\author{
    Rui Zhao$^{1,2,3}$, \ Xuewen Zhong$^{1,2,3}$, \ Xiaoyun Zheng$^{1,2,3}$, \\ {\bf Jinsong Su$^{1,2}$} \and {\bf Yidong Chen$^{1,2,3}$}\thanks{Corresponding Author.}\\
    $^1$School of Informatics, Xiamen University, China \\
    $^2$Key Lab of Digital Protection and Intelligent Processing of Intangible \\ Cultural Heritage of Fujian-Taiwan (XMU), Ministry of Culture and Tourism, China \\
    $^3$National Language Resources Monitoring and \\ Research Center for Education and Teaching Media, Xiamen University, China \\
    \texttt{zhsqzr@stu.xmu.edu.cn \quad ydchen@xmu.edu.cn} \\
}

\begin{document}
\maketitle

\begin{abstract}

Sign language research has achieved significant progress due to the advances in large language models (LLMs). However, the intrinsic ability of LLMs to understand sign language, especially in multimodal contexts, remains underexplored. 
To address this limitation, we introduce \textbf{CNSL-bench}, the first comprehensive \textbf{C}hinese \textbf{N}ational \textbf{S}ign \textbf{L}anguage \textbf{bench}mark designed for evaluating multimodal large language models (MLLMs) in sign language understanding. 
The proposed CNSL-bench is characterized by: 
1) Authoritative grounding, as it is anchored to the officially standardized \textit{National Common Sign Language Dictionary}, mitigating ambiguity from regional or non-canonical variants and ensuring consistent semantic definitions; 
2) Multimodal coverage, providing aligned textual descriptions, illustrative images, and sign language videos; 
and 3) Articulatory diversity, supporting fine-grained analysis across key manual articulatory forms, including air-writing, finger-spelling, and the Chinese manual-alphabet.
Using CNSL-bench, we extensively evaluate 21 open-source and proprietary up-to-date MLLMs. 
Our results reveal that, despite recent advances in multimodal modeling, current MLLMs remain substantially inferior to human performance, exhibiting systematic disparities across input modalities and manual articulatory forms.
Additional diagnostic analyses suggest that several performance limitations persist beyond improvements in reasoning and that instruction-following robustness varies substantially across models.

%


\end{abstract}

\begin{figure}[t]
\centering
\includegraphics[width=0.95 \columnwidth]{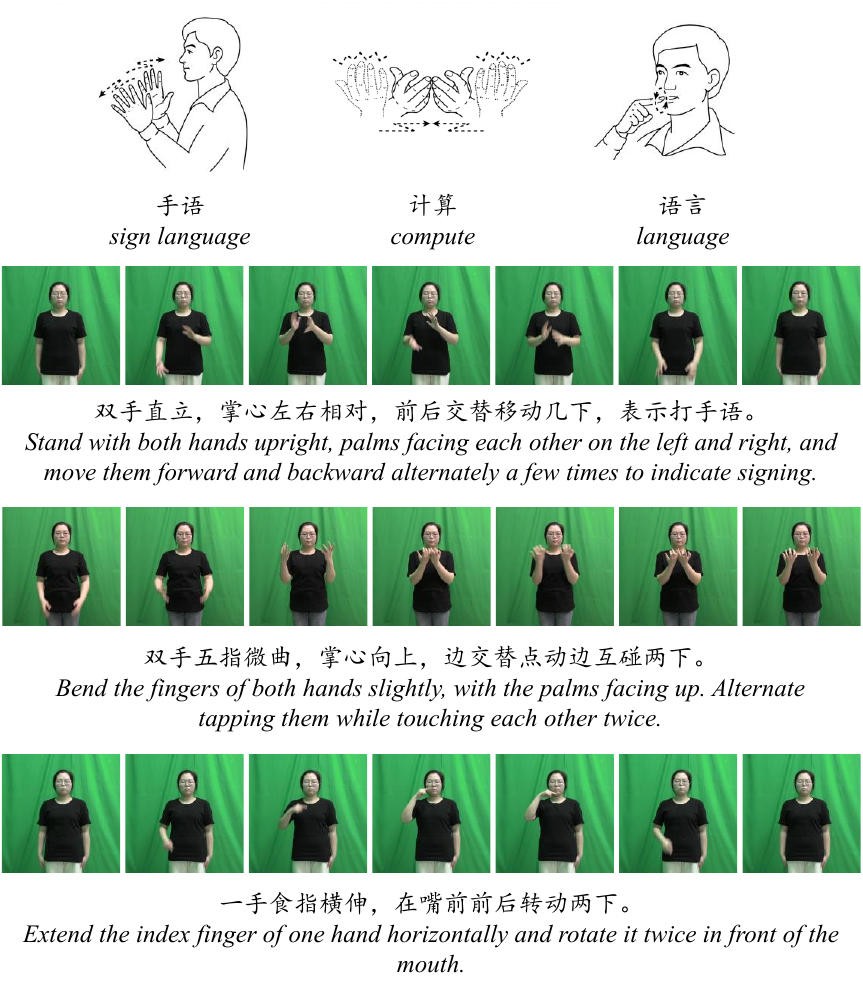}
\caption{Examples from CNSL-bench showing the aligned textual description, illustrative image, and corresponding sign language video for each sign entry.}
\label{fig_introduction}
\end{figure}

\section{Introduction}

Sign language plays a central role in communication for many people with hearing impairment and has consequently attracted sustained attention from the research community over the past decades. More recently, advances in large language models (LLMs) have further stimulated progress in automatic sign language understanding, primarily within specific tasks, e.g., sign language translation, where LLMs are incorporated as semantic augmentation modules or enhanced text decoders for improved performance~\cite{wongSign2GPTLeveragingLarge2023, gongLLMsAreGood2024a, chenFactorizedLearningAssisted2024a, guoBridgingSignSpoken2025a, kimLeveragingPowerMLLMs2025b, liuSCOPESignLanguage2025, jangLostTranslationFound2025a, asasiGlossHandCentricFramework2025a, raoRVLFReinforcingVisionLanguage2025, hwangEfficientGlossFreeSign2025}.

Despite encouraging task-level improvements, existing approaches predominantly embed LLMs into downstream pipelines or datasets, leaving the intrinsic ability of these models to understand sign language largely unexamined. This limitation becomes even more pronounced in the context of multimodal large language models (MLLMs), which have demonstrated strong visual-language capabilities over images and videos~\cite{liuVisualInstructionTuning2023, liu2024llavanext, zhangMultimodalChainofThoughtReasoning2024, shenVLMR1StableGeneralizable2025}. 
Crucially, sign language possesses the full range of fundamental linguistic properties and is inherently multimodal, with meaning expressed through the coordinated use of linguistically grounded manual articulatory cues such as air writing, finger-spelling, and the manual-alphabet~\cite{shiFingerspellingDetectionAmerican2021, yinIncludingSignedLanguages2021, desaiSystemicBiasesSign2024, atwellStudyingMitigatingBiases2024}. 
While this rich expressiveness poses fundamental challenges for automatic sign language understanding, it also raises an open question: to what extent can current MLLMs genuinely comprehend sign language by capturing linguistically grounded structure and semantic meanings rather than relying solely on visual correlations?

\begin{figure*}[ht]
\centering
\includegraphics[width=1.0 \linewidth]{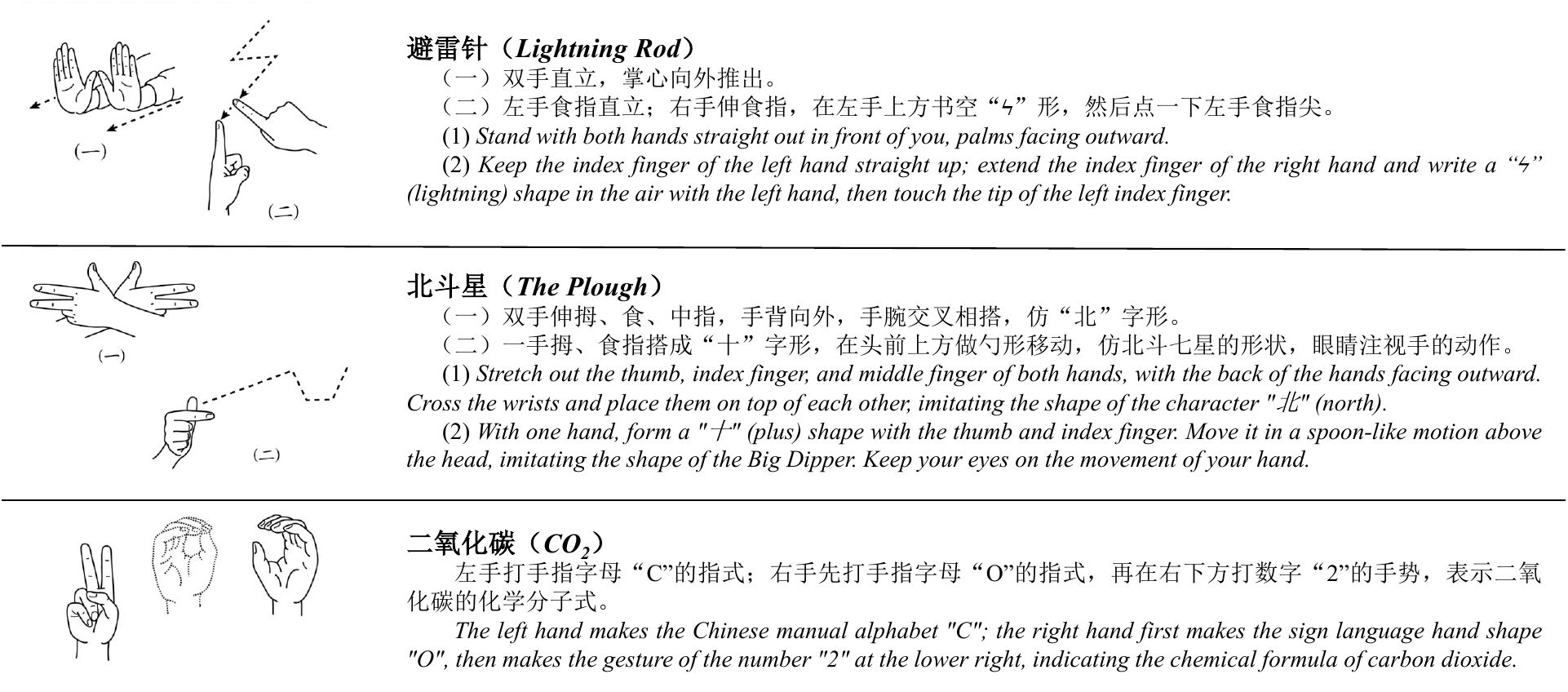}
\caption{Examples illustrating the three categories of sign articulation: air-writing (top), finger-spelling (middle), and manual-alphabet (bottom).}
\label{fig_subset}
\end{figure*}

To answer this question, we introduce \textbf{CNSL-bench}, the first comprehensive \textbf{C}hinese \textbf{N}ational \textbf{S}ign \textbf{L}anguage \textbf{bench}mark designed for evaluating MLLMs in sign language understanding (\textbf{\S}\ref{sec_cnsl_construct}).
CNSL-bench is constructed upon the \emph{National Common Sign Language Dictionary}, an officially standardized lexical resource for Chinese national sign language. 
This authoritative grounding provides a canonical semantic reference, reducing ambiguity from regional or variants and enabling consistent, controlled evaluation of sign language understanding~\cite{cnsl_common, cnsl_full}. 
To enable multimodal evaluation, we further align these dictionary entries with a large-scale video dataset covering isolated Chinese national sign language~\cite{jinLargeDatasetCovering2025}, resulting in a unified benchmark that provides broad multimodal coverage, where each sign entry is represented by aligned \emph{textual description}, \emph{illustrative image}, and \emph{sign language video}, as exemplified in Figure~\ref{fig_introduction}. 
As a consequence, CNSL-bench comprises 20,121 questions spanning text, image, and video. 
Furthermore, the benchmark explicitly supports manual articulatory diversity, covering three representative categories of signs, including air-writing, finger-spelling, and manual-alphabet, as shown in Figure~\ref{fig_subset}. These categories are treated as dedicated evaluation subsets to enable fine-grained analysis. 
Supported by high-quality resources, carefully designed evaluation protocols, and human assessment involving the Deaf\footnote{We follow the recognized convention of using the upper-cased word Deaf to refer to the community of sign language users~\cite{woodwardImplicationsSociolinguisticResearch1972}} community, CNSL-bench serves as a reliable and comprehensive benchmark for systematically diagnosing the intrinsic sign language understanding capabilities of modern MLLMs.

With CNSL-bench, we benchmark a wide range of open- and closed-source up-to-date MLLMs, offering a systematic empirical assessment of their sign language understanding capabilities (\textbf{\S}~\ref{sec_main_results}). 
Our main results reveal several patterns: 
\textbf{(a)} Despite recent advances in multimodal modeling, current MLLMs remain substantially inferior to human performance in sign language understanding. 
\textbf{(b)}  A pronounced modality-dependent performance imbalance is observed, with models exhibiting substantially weaker performance on visual inputs compared to text. 
\textbf{(c)} MLLMs demonstrate uneven comprehension across different manual articulatory forms, achieving relatively stronger performance on finger-spelling than on air-writing and the specific manual-alphabet. 
\textbf{(d)} The performance gap between open-source and proprietary MLLMs is rapidly narrowing, with several open-source models achieving performance comparable to lightweight commercial systems. 
Beyond the main results, additional diagnostic analyses are conducted to further investigate the intrinsic sign language understanding capabilities of current MLLMs (\textbf{\S}~\ref{sec_cot}). Test-time scaling via explicit reasoning mechanisms is examined as a diagnostic tool, revealing heterogeneous gains across models and modalities while failing to fundamentally resolve the observed limitations. Prompt token effects and instruction-following robustness are further considered as complementary diagnostic factors, providing additional insight into the behavioral characteristics of current MLLMs.


%

In summary, the contributions of this work are as follows: 

(1) We introduce \textbf{CNSL-bench}, a multimodal, sign language-centric benchmark for evaluating sign language understanding in MLLMs. 

(2) We present a comprehensive evaluation of up-to-date MLLMs on the proposed CNSL-bench.

(3) Through extensive experiments and analyses, we identify persistent challenges in current MLLMs' sign language understanding capabilities.

We hope that CNSL-bench will serve as a diagnostic foundation and a reference resource for future research toward more robust, reliable, and human-aligned MLLMs in sign language understanding. Data and code are available at \url{https://github.com/rzhao-zhsq/CNSL-bench}.

\section{CNSL-bench}\label{sec_cnsl_bench}


%

Sign language understanding requires unified modeling of visual, temporal, and fine-grained linguistic information across heterogeneous modalities. To enable a systematic and controlled evaluation of such capabilities in MLLMs, we construct CNSL-bench, a Chinese National Sign Language benchmark grounded in standardized lexical resources and aligned multimodal representations. CNSL-bench is designed to isolate intrinsic sign language understanding from downstream task-specific factors, providing a consistent evaluation framework across text, image, and video inputs.

\subsection{Benchmark Construction}\label{sec_cnsl_construct}

\paragraph{Benchmark Principle.}


CNSL-bench is constructed following three core principles: 
1) it is grounded in a standardized lexical foundation. To avoid ambiguity introduced by regional or non-canonical sign variants, the benchmark is anchored to the officially standardized National Common Sign Language Dictionary, ensuring consistent semantic definitions across all samples; 
2) it emphasizes aligned multimodal coverage. Each sign entry is systematically represented in text, image, and video formats, enabling controlled evaluation of cross-modal sign language understanding; 
and 3) the benchmark explicitly incorporates diverse manual articulatory forms, including air-writing, finger-spelling, and specific manual-alphabet signs, supporting fine-grained analysis of linguistically distinct forms in sign language.

\begin{figure*}[ht]
\centering
    \begin{subfigure}{0.48\linewidth}
    \centering
    \includegraphics[width=\linewidth]{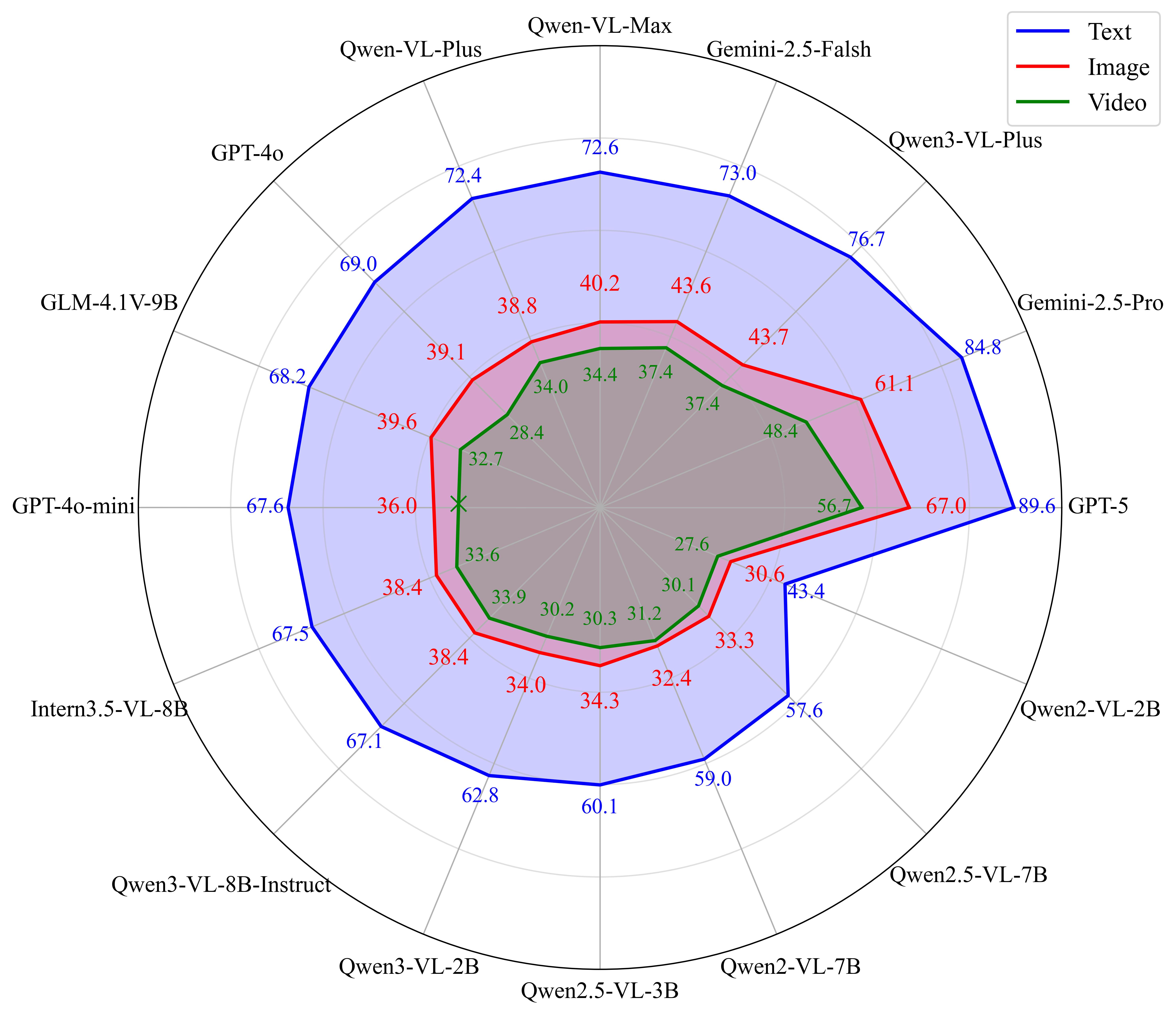}
    \caption{Overall performance on CNSL-bench.}
    \label{fig_radar1}
    \end{subfigure}
\hfill
    \begin{subfigure}{0.48\linewidth}
    \centering
    \includegraphics[width=\linewidth]{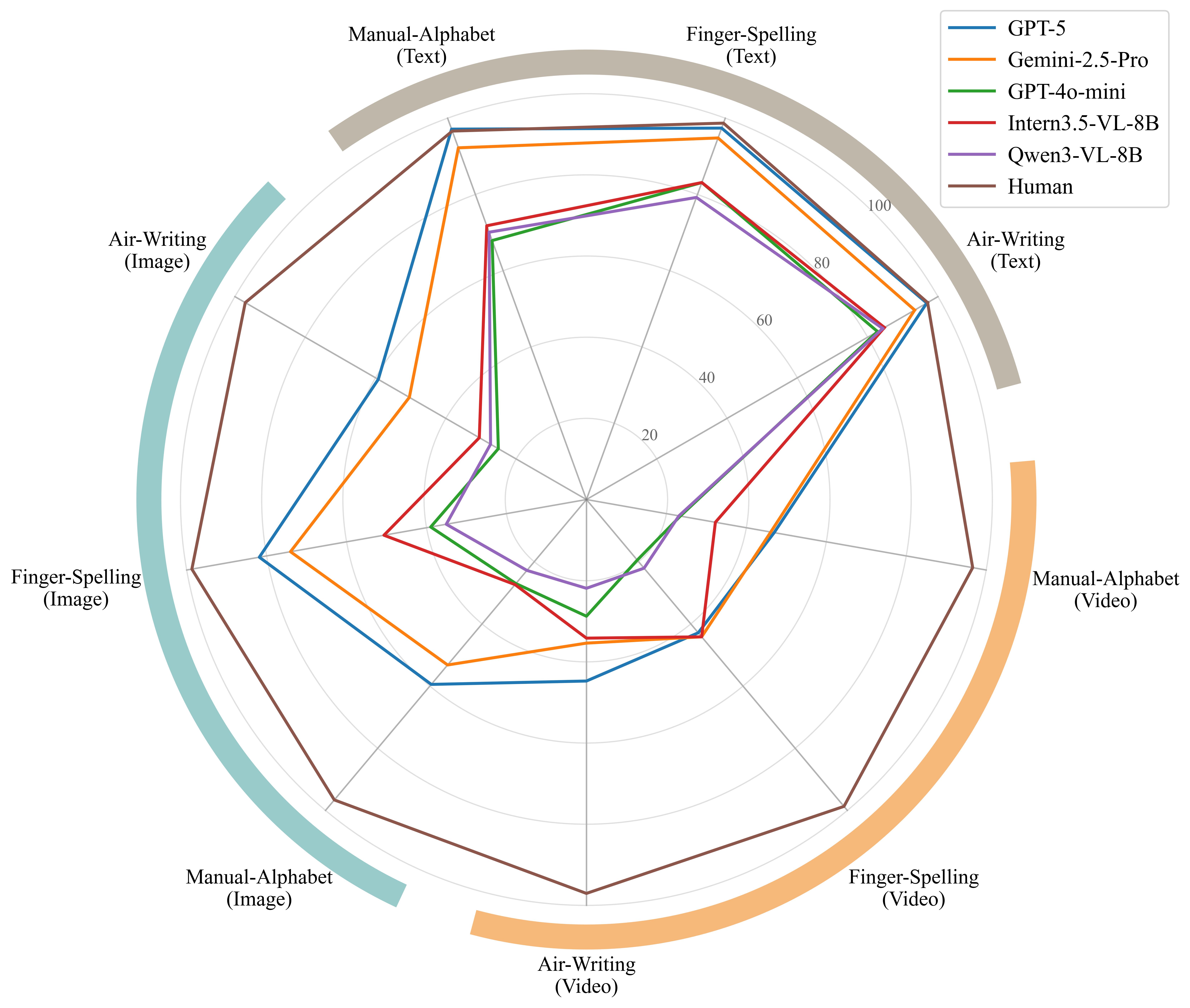}
    \caption{Performance of subsets on CNSL-bench.}
    \label{fig_radar2}
    \end{subfigure}
\caption{A concise summary of the MLLMs’ performance on CNSL-bench. (a) provides a comparison detailing the overall performance of the 16 selected MLLMs, including both open-source and closed-source models. (b) illustrates a radar chart that outlines the performance of humans and MLLMs on three subsets (air-writing, finger-spelling, and manual-alphabet) within CNSL-bench.
}
\label{fig_radar}
\end{figure*}

\begin{table*}[ht]

\resizebox{1.0 \linewidth}{!} {
\begin{tabular}{@{}lcccccccccccccccc@{}}
\toprule
\multirow{2}{*}{Model} & \multicolumn{4}{c}{Text} & \multicolumn{4}{c}{Image} & \multicolumn{4}{c}{Video$^{'2~fps}$} & \multicolumn{4}{c}{Video$^{'10~fps}$} \\ 
\cmidrule(r){2-5} \cmidrule(lr){6-9} \cmidrule(lr){10-13} \cmidrule(l){14-17}
 & AW & FS & MA & All & AW & FS & MA & All & AW & FS & MA & All & AW & FS & MA & All \\ 
\midrule
\multicolumn{17}{c}{\cellcolor[HTML]{EFEFEF}\textit{Open\&Close -source Image MLLMs}} \\ 
\midrule
LLaVA-NeXT-7B & 0.74 & 2.60 & 0.68 & 1.68 & 17.20 & 14.29 & 17.57 & 17.52 & 15.72 & 12.99 & 17.74 & 18.67 & - & - & - & - \\
Qwen-VL-Plus & 87.71 & 84.42 & 77.97 & 72.41 & 25.37 & 28.57 & 25.89 & 38.83 & 17.69 & 27.27 & 21.83 & 32.06 & 21.38 & 22.08 & 23.99 & 33.95 \\
Qwen-VL-Max & 85.50 & 83.12 & 77.80 & 72.62 & 24.88 & 33.77 & 24.87 & 40.18 & 20.64 & 23.38 & 22.00 & 31.01 & 24.08 & 31.17 & 23.99 & 34.43 \\ 

\midrule
\multicolumn{17}{c}{\cellcolor[HTML]{EFEFEF}\textit{Open-Source MLLMs}} \\ 
\midrule

Qwen2-VL-2B & 47.42 & 51.95 & 41.55 & 43.36 & 19.41 & 19.48 & 23.65 & 30.62 & {\ul 20.39} & \textbf{27.27} & \textbf{22.80} & 27.23 & 19.16 & 18.18 & \textbf{23.65} & 27.58 \\
Qwen2.5-VL-3B & 69.04 & 72.73 & 55.41 & 60.07 & {\ul 25.80} & 27.27 & {\ul 23.82} & {\ul 34.26} & \textbf{20.64} & {\ul 24.68} & 17.91 & \textbf{28.36} & \textbf{23.59} & \textbf{28.57} & 20.61 & {\ul 30.34} \\
Intern-VL-3.5-2B & 68.06 & {\ul 79.22} & \textbf{60.64} & 58.68 & 19.66 & 28.57 & \textbf{24.49} & 33.37 & 2.95 & 6.49 & 4.39 & 4.29 & 4.42 & 6.49 & 4.39 & 4.50 \\
Qwen3-VL-2B-Instruct & \textbf{73.71} & \textbf{83.12} & {\ul 58.61} & \textbf{62.83} & 21.62 & {\ul 38.96} & 21.62 & 34.01 & 19.66 & 18.18 & {\ul 21.79} & {\ul 27.78} & 21.38 & 19.48 & {\ul 23.14} & 30.18 \\
Qwen3-VL-2B & 66.58 & 74.03 & 52.70 & 57.97 & 23.59 & 27.27 & 20.78 & 32.80 & {\ul 20.39} & 16.88 & 17.74 & 26.38 & {\ul 22.36} & 23.38 & 18.07 & 28.95 \\
Qwen3-VL-2B~\ding{45} & {\ul 72.97} & 75.32 & 56.25 & {\ul 61.29} & \textbf{26.54} & \textbf{41.56} & 21.28 & \textbf{35.37} & 16.71 & 15.58 & 19.76 & 27.76 & {\ul 22.36} & {\ul 27.27} & 20.95 & \textbf{30.68} \\

\cmidrule(lr){2-5} \cmidrule(lr){6-9} \cmidrule(lr){10-13} \cmidrule(l){14-17}

LLaVA-NeXT-Video-7B & 1.72 & 2.60 & 0.84 & 1.34 & 8.85 & 10.39 & 12.16 & 12.94 & 14.99 & 18.18 & 15.20 & 15.43 & 13.76 & 14.29 & 15.71 & 15.91 \\
Qwen2-VL-7B & 68.30 & 75.32 & 56.93 & 58.98 & 26.78 & 31.17 & 25.17 & 32.44 & {\ul 22.60} & {\ul 29.87} & 21.45 & 29.24 & 23.10 & {\ul 27.27} & 23.14 & 31.19 \\
Qwen2.5-VL-7B & 65.60 & 70.13 & 55.41 & 57.61 & 26.29 & 27.27 & 24.83 & 33.32 & 21.38 & 24.68 & 20.95 & 29.10 & 22.11 & 23.38 & 21.79 & 30.15 \\
GLM-4.1V-9B~\ding{45} & 80.34 & \textbf{84.42} & 69.59 & {\ul 68.24} & {\ul 28.50} & \textbf{55.84} & \textbf{28.38} & \textbf{39.62} & 20.39 & 23.38 & 19.76 & 28.03 & 21.87 & 24.68 & 21.62 & 29.75 \\
Intern-VL-3.5-8B & {\ul 84.77} & {\ul 83.12} & {\ul 71.79} & 67.53 & \textbf{30.47} & {\ul 50.65} & {\ul 27.36} & 38.36 & \textbf{34.15} & \textbf{44.16} & \textbf{32.26} & \textbf{32.26} & \textbf{31.70} & \textbf{36.36} & \textbf{29.39} & {\ul 33.59} \\
Qwen3-VL-8B-Instruct & 84.28 & 79.22 & 70.10 & 67.06 & 27.27 & 35.06 & 22.80 & {\ul 38.39} & 21.87 & 22.08 & {\ul 23.14} & {\ul 30.94} & {\ul 25.80} & 24.68 & {\ul 28.38} & \textbf{33.89} \\
Qwen3-VL-8B & 81.82 & 80.52 & 67.40 & 64.65 & 26.04 & 28.57 & 25.68 & 36.34 & {\ul 22.60} & 15.58 & 19.93 & 28.51 & 20.39 & 20.78 & 22.13 & 30.42 \\
Qwen3-VL-8B~\ding{45} & \textbf{87.22} & {\ul 83.12} & \textbf{78.89} & \textbf{70.61} & 27.03 & 29.87 & 23.65 & 37.56 & 22.11 & 18.18 & 19.93 & 29.88 & 24.57 & 25.97 & 25.00 & 33.00 \\ 


\midrule
\multicolumn{17}{c}{\cellcolor[HTML]{EFEFEF}\textit{Closed-Source MLLMs}} \\ 
\midrule

GPT-4o-mini & 82.80 & 83.12 & 67.91 & 67.57 & 25.06 & 38.96 & 27.03 & 35.99 & 28.75 & 19.48 & 23.48 & 27.30 & - & - & - & - \\
GPT-4o & 82.06 & 88.31 & 72.13 & 69.03 & 32.43 & 51.95 & 29.73 & 39.07 & 27.76 & 20.78 & 26.01 & 31.26 & 25.31 & 23.38 & 21.96 & 28.43 \\
Qwen3-VL-Plus & 90.91 & 88.31 & 78.14 & 76.68 & 28.57 & 32.47 & 27.07 & 43.69 & 18.43 & 20.78 & 25.21 & 33.74 & 24.57 & 24.68 & 28.21 & 37.37 \\
Qwen3-VL-Plus~\ding{45} & 92.38 & 89.61 & 84.24 & 76.22 & 31.77 & 38.96 & 29.95 & 42.41 & 26.85 & 18.18 & 25.93 & 35.34 & 24.57 & 18.18 & 25.17 & 36.92 \\
Gemini-2.5-Falsh & 85.01 & 88.31 & 72.64 & 73.04 & 30.47 & 31.93 & 45.58 & 43.57 & 28.26 & 23.38 & 30.07 & 36.62 & 26.04 & 24.68 & 29.39 & 37.44 \\
Gemini-2.5-Falsh~\ding{45} & 92.38 & 93.51 & 83.11 & 79.95 & 34.64 & 51.95 & 35.47 & 51.62 & 30.47 & 33.77 & 31.93 & 42.28 & 32.43 & 32.47 & 36.32 & 42.63 \\
Gemini-2.5-Pro~\ding{45} & {\ul 93.37} & {\ul 94.81} & {\ul 92.23} & {\ul 84.79} & {\ul 50.37} & {\ul 74.03} & {\ul 53.21} & {\ul 61.13} & {\ul 35.38} & \textbf{44.16} & {\ul 46.11} & {\ul 48.32} & {\ul 36.61} & {\ul 35.06} & {\ul 39.02} & {\ul 48.35} \\
GPT-5~\ding{45} & \textbf{96.81} & \textbf{97.40} & \textbf{97.13} & \textbf{89.64} & \textbf{59.21} & \textbf{81.82} & \textbf{59.46} & \textbf{66.96} & \textbf{44.72} & {\ul 42.86} & \textbf{46.96} & \textbf{53.42} & \textbf{46.93} & \textbf{53.25} & \textbf{53.55} & \textbf{56.72} \\

\midrule
Random & 25.35 & 27.27 & 24.69 & 25.23 & 24.57 & 24.67 & 24.49 & 24.73 & 25.79 & 23.98 & 24.92 & 25.03 & 24.57 & 24.67 & 24.76 & 25.04 \\
Human & 98.77 & 97.40 & 96.96 & 96.93 & 98.77 & 98.70 & 98.31 & 97.39 & 99.26 & 98.70 & 97.47 & 97.39 & 99.26 & 98.70 & 97.47 & 97.39 \\ 
\bottomrule
\end{tabular}
}
\caption{Performance of MLLMs on CNSL-bench. AW, FS, and MA indicate air-writing, finger-spelling, and manual-alphabet. 
\ding{45} denotes inference with slow thinking. 
The best result is \textbf{bolded}, and the second is {\ul underlined}.
}
\label{table_main_results}
\centering
\end{table*}

\paragraph{Data Collection.}

CNSL-bench is constructed by grounding all entries in officially standardized Chinese national sign language resources and aligning them with multimodal representations. The core lexical inventory is derived from the \textit{National Common Sign Language Dictionary}~\cite{cnsl_full}, which is built upon the \textit{Lexicon of Common Expressions in Chinese National Sign Language} jointly issued by the Ministry of Education, the State Language Commission, and the China Disabled Persons’ Federation~\cite{cnsl_common}. 
Together, these standards define normative sign realizations that are widely used and stable in education and daily communication. This grounding ensures lexical consistency and reduces regional or informal variation. 
To ensure that each benchmark item maps to a unique sign entry, we perform systematic sign-level preprocessing to handle dictionary cases where (i) distinct entries share identical hand motions, (ii) identical entries are associated with different meanings or articulations (e.g., ``seat belt'' for car and airplane), or (iii) the same meaning is realized by distinct hand motions.
For each entry, we retain the accompanying textual description and illustrative image, which explain the sign entry for educational and communication purposes.
Furthermore, each unique sign entry is aligned with an additional sign language video to reflect real-world usage. The video samples are sourced from a large-scale Chinese national sign language dataset~\cite{jinLargeDatasetCovering2025}, yielding a unified multimodal representation for each sign entry.
%

In addition, CNSL-bench explicitly incorporates specific manual articulatory forms, including air-writing, finger-spelling, and the Chinese manual-alphabet. 
As exemplified in Figure~\ref{fig_subset}, air-writing refers to tracing graphic forms in the air, and the traced content may correspond to strokes, symbol-like shapes, or a partial character. 
In our taxonomy, finger-spelling refers to using one or both hands to depict or indicate Chinese character-form structure, prioritizing graphic cues such as outlines, components, and structural patterns over strictly sequential, letter-by-letter spelling.
The manual-alphabet follows the \textit{Chinese Manual Alphabet}~\cite{cnsl_alphabet}, which maps conventionalized finger configurations to individual Chinese Pinyin letters. These letters can be combined to spell Mandarin, forming lexical signs, and functioning as morphemic components within signs under the \textit{Scheme of the Chinese Phonetic Alphabet}~\cite{scheme_of_chinese_phonetic_alphabet}. 

The alignment between text, image, and video is detailed in Appendix~\ref{sec_app_cnsl_alignment}, and the complete manual alphabets in the \textit{Chinese Manual Alphabet} are listed in Figure~\ref{fig_cnsl_alphabet} (Appendix~\ref{sec_app_cnsl_alphabet}).
%



\paragraph{Task Definition.} 
For each sign entry, CNSL-bench provides an aligned textual description, an illustrative image, and a sign language video. We formulate the task as a four-way multiple-choice evaluation, where each instance is instantiated with a single input modality. This closed-form design enables controlled and scalable evaluation, as current state-of-the-art MLLMs remain highly unreliable in open-ended sign language understanding. We further examine different option construction strategies and observe that semantics-based distractors lead to slightly weaker model performance, while yielding conclusions consistent with random sampling. To simplify the benchmark design and facilitate easy reproduction, we therefore adopt random option sampling for robustness and consistency. 
Detailed task specifications and comparative analyses of option construction strategies, including an open-ended case and distractor design, are provided in Appendix~\ref{sec_app_task}.

\subsection{Benchmark Statistics}

CNSL-bench is grounded in the \textit{National Common Sign Language Dictionary}, which contains 8,214 sign glosses. However, it includes cases where different glosses share identical hand motions, identical glosses correspond to different meanings and articulations, or the same meaning is realized through multiple distinct hand motions. 
After processing at the sign-entry level, CNSL-bench retains 6,707 unique sign entries, yielding 20,121 evaluation instances across three modalities (text, image, and video).
Among the 6,707 sign entries, we manually identify 407 entries containing air-writing, 77 containing finger-spelling, and 592 involving specific manual-alphabet, enabling dedicated subset evaluation for sign-linguistical sensitive analysis.
Moreover, a sign entry may comprise multiple atomic gestures due to sequential articulation or multi-part realizations, with up to 7 gestures in the most complex cases. 
The detailed statistics and breakdowns are provided in Appendix~\ref{sec_app_statistics}.

Overall, CNSL-bench comprises 20,121 questions spanning text, image, and video modalities, with each question grounded in a standardized lexical entry and aligned multimodal evidence. This construction supports systematic and fine-grained evaluation of sign language understanding under a unified benchmark setting.

\begin{table*}
\centering
\resizebox{1.0 \linewidth}{!} {
\begin{tabular}{@{}lcccccccccccccccc@{}}
\toprule
\multirow{2}{*}{Model} & \multicolumn{4}{c}{Text} & \multicolumn{4}{c}{Image} & \multicolumn{4}{c}{Video$^{'2~fps}$} & \multicolumn{4}{c}{Video$^{'10~fps}$} \\ 
\cmidrule(r){2-5} \cmidrule(lr){6-9} \cmidrule(lr){10-13} \cmidrule(l){14-17}
 & AW & FS & MA & All & AW & FS & MA & All & AW & FS & MA & All & AW & FS & MA & All \\ 
\midrule
\multicolumn{17}{c}{\cellcolor[HTML]{EFEFEF}\textit{Fast Thinking}} \\ 
\midrule
Qwen3-VL-2B & 66.58 & 74.03 & 52.70 & 57.97 & 23.59 & 27.27 & 20.78 & 32.80 & 20.39 & 16.88 & 17.74 & 26.38 & 22.36 & 23.38 & 18.07 & 28.95 \\
Qwen3-VL-8B & 81.82 & 80.52 & 67.40 & 64.65 & 26.04 & 28.57 & 25.68 & 36.34 & 22.60 & 15.58 & 19.93 & 28.51 & 20.39 & 20.78 & 22.13 & 30.42 \\
Qwen3-VL-Plus & 90.91 & 88.31 & 78.14 & 76.68 & 28.57 & 32.47 & 27.07 & 43.69 & 18.43 & 20.78 & 25.21 & 33.74 & 24.57 & 24.68 & 28.21 & 37.37 \\
Gemini-2.5-Flash & 85.01 & 88.31 & 72.64 & 73.04 & 30.47 & 31.93 & 45.58 & 43.57 & 28.26 & 23.38 & 30.07 & 36.62 & 26.04 & 24.68 & 29.39 & 37.44 \\
\midrule
\multicolumn{17}{c}{\cellcolor[HTML]{EFEFEF}\textit{Slow Thinking}} \\ 
\midrule
Qwen3-VL-2B & 72.97 & 75.32 & 56.25 & 61.29 & 26.54 & 41.56 & 21.28 & 35.37 & 16.71 & 15.58 & 19.76 & 27.76 & 22.36 & 27.27 & 20.95 & 30.68 \\
Qwen3-VL-8B & 87.22 & 83.12 & 78.89 & 70.61 & 27.03 & 29.87 & 23.65 & 37.56 & 22.11 & 18.18 & 19.93 & 29.88 & 24.57 & 25.97 & 25.00 & 33.00 \\
Qwen3-VL-Plus & 92.38 & 89.61 & 84.24 & 76.22 & 31.77 & 38.96 & 29.95 & 42.41 & 26.85 & 18.18 & 25.93 & 35.34 & 24.57 & 18.18 & 25.17 & 36.92 \\
Gemini-2.5-Flash & 92.38 & 93.51 & 83.11 & 79.95 & 34.64 & 51.95 & 35.47 & 51.62 & 30.47 & 33.77 & 31.93 & 42.28 & 32.43 & 32.47 & 36.32 & 42.63 \\
\cmidrule(lr){2-5} \cmidrule(lr){6-9} \cmidrule(lr){10-13} \cmidrule(l){14-17}
Gemini-2.5-Pro (L) & 91.15 & 96.10 & 86.15 & 81.32 & 46.68 & 72.73 & 52.20 & 58.09 & 37.59 & 42.86 & 44.76 & 48.83 & 34.64 & 42.86 & 41.72 & 47.96 \\
Gemini-2.5-Pro (M) & 93.37 & 94.81 & 92.23 & 84.79 & 50.37 & 74.03 & 53.21 & 61.13 & 35.38 & 44.16 & 46.11 & 48.32 & 36.61 & 35.06 & 39.02 & 48.35 \\
Gemini-2.5-Pro (H) & 92.63 & 94.81 & 90.71 & 84.84 & 49.63 & 74.03 & 54.05 & 61.92 & 36.61 & 38.96 & 42.40 & 48.17 & 38.57 & 36.36 & 45.10 & 48.59 \\
GPT-5 (L) & 97.05 & 97.40 & 95.61 & 88.94 & 55.53 & 74.03 & 60.64 & 66.77 & 40.54 & 38.96 & 46.45 & 51.89 & 41.03 & 48.05 & 45.10 & 53.13 \\
GPT-5 (M) & 96.81 & 97.40 & 97.13 & 89.64 & 59.21 & 81.82 & 59.46 & 66.96 & 44.72 & 42.86 & 46.96 & 53.42 & 46.93 & 53.25 & 53.55 & 56.72 \\
GPT-5 (H) & 97.05 & 98.70 & 96.62 & 89.95 & 63.14 & 83.12 & 59.46 & 68.34 & 42.51 & 37.66 & 48.99 & 53.09 & 44.58 & 38.96 & 46.45 & 54.01 \\
\bottomrule
\end{tabular}
}
\caption{Numerical results with test-time scaling on reasoning models. L, M, H: low, medium, and high reasoning effort on the process of thinking before generating an answer.}
\label{table_cot_results}
\end{table*}

\begin{figure*}[h]
\centering
\includegraphics[width=1.0 \linewidth]{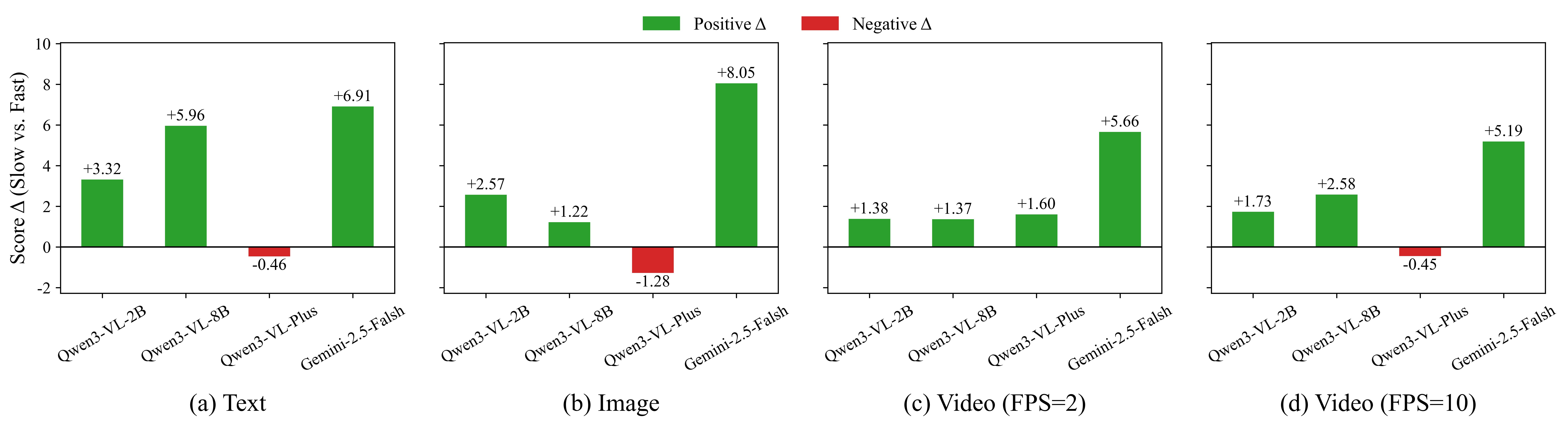}
\caption{CoT gains across models and multimodal subsets.}
\label{fig_slow_fast_1}
\end{figure*}

\begin{figure*}[h]
\centering
\includegraphics[width=1.0 \linewidth]{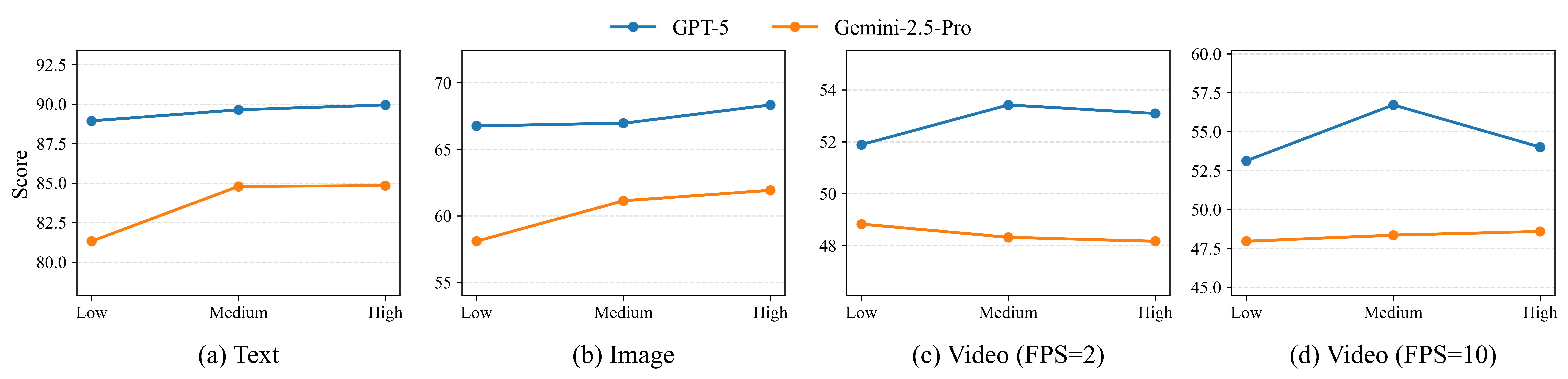}
\caption{
Top-tier reasoning models, i.e., GPT-5 and Gemini-2.5-Pro, exhibit a boundary effect in test-time scaling as reasoning effort increases from low to high, especially in video input settings.
}
\label{fig_slow_fast_2}
\end{figure*}

\section{Experiments}

\subsection{Experimental Settings}
We evaluate 21 up-to-date MLLMs, spanning open-source families (LLaVA-NeXT, Qwen-VL, InternVL-3.5, GLM-4.1V) and closed-source models (Qwen-Plus/Max, Gemini-2.5, GPT-4/5). 
Detailed inference configurations and the human evaluation protocol are provided in Appendix~\ref{sec_app_settings}.

\subsection{Main Results}\label{sec_main_results}


The overall performance and subcategory comparisons (human vs. representative MLLMs) on CNSL-Bench can be quickly glanced at Figure~\ref{fig_radar}. 
Table~\ref{table_main_results} details the overall performance of a wide range of MLLMs on CNSL-bench across modalities (i.e., text, image, and sign video) and manual articulatory forms (i.e., AW, FS, and MA).

\paragraph{Human--MLLMs performance gap.}
Despite recent progress, current MLLMs remain markedly inferior to human-level sign language understanding across all modalities. The strongest proprietary model, GPT-5, achieves overall accuracies of 89.64\%, 66.96\%, and 56.72\% on text, image, and sign language video understanding, respectively. Although GPT-5 substantially outperforms other advanced models, a clear and persistent gap remains when compared to human performance, which consistently reaches approximately 97\% across all three modalities. This discrepancy highlights the fundamental difficulty of achieving robust, human-level comprehension of sign language, particularly in visually grounded and temporally complex settings.

\paragraph{Modality--dependent performance imbalance.}
A pronounced modality imbalance is consistently observed in MLLMs’ sign language understanding. Across nearly all models, performance is highest for textual descriptions, while accuracy drops substantially for illustrative images and further degrades for sign language videos. This trend holds for both open-source and closed-source systems and becomes more pronounced in the video setting. 
These results indicate that, although text-centric language modeling is relatively well developed, robust visual grounding and temporal modeling remain major challenges for current MLLMs. 

\paragraph{Uneven understanding across manual articulatory forms.}
MLLMs exhibit uneven comprehension across different manual articulatory forms. As shown in Table~\ref{table_main_results}, models consistently achieve stronger performance on finger-spelling than on air-writing and manual-alphabet across all modalities. This disparity suggests that current models handle more discrete and character-like sign components more reliably, while continuous, shape-intensive, or motion-dependent articulations remain difficult. 

\paragraph{Narrowing gap between open- and closed-source MLLMs.}
The performance gap between open-source and proprietary MLLMs is rapidly narrowing. Several small-scale open-source models, including GLM-4.1V-9B, InternVL-3.5-8B, and Qwen3-VL-8B, attain performance comparable to lightweight commercial systems such as GPT-4o-mini and Gemini-2.5-Flash across multiple modalities. 
These models even surpass proprietary counterparts on specific subsets, underscoring the rapid advancement and increasing competitiveness of open-source MLLMs in sign language understanding. 
Nevertheless, despite these encouraging trends, open-source models still need to make continued progress on challenging subsets and to match higher-capacity proprietary systems, particularly under multimodal settings.

%

\begin{figure}[t]
\centering
\includegraphics[width=1.0\linewidth]{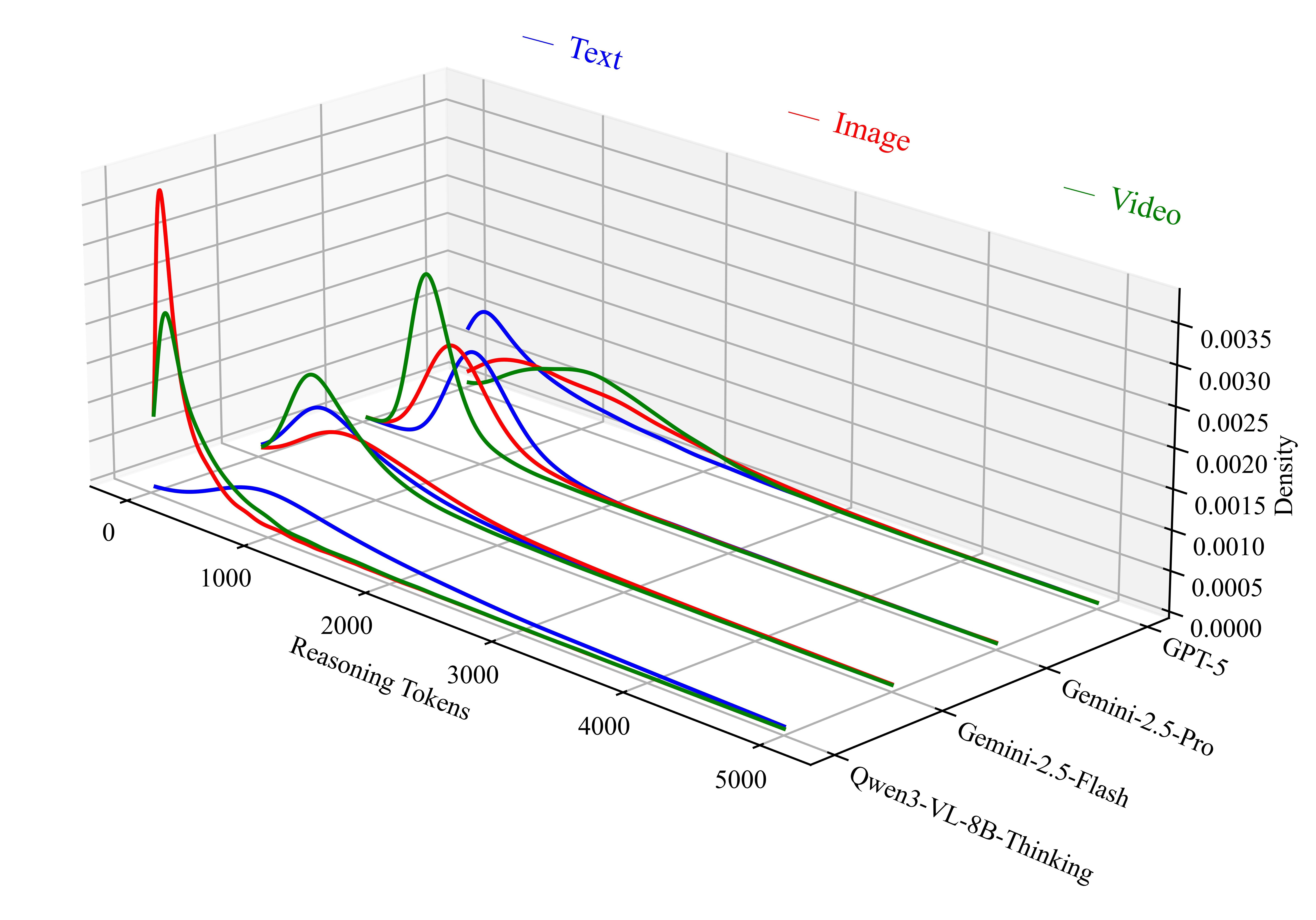}
\caption{
KDE plot of reasoning tokens. Videos are at 10 frames per second. Zoom in for better visualization.
}
\label{fig_kde}
\end{figure}

\begin{figure*}[t]
\centering
\includegraphics[width=1.0 \linewidth]{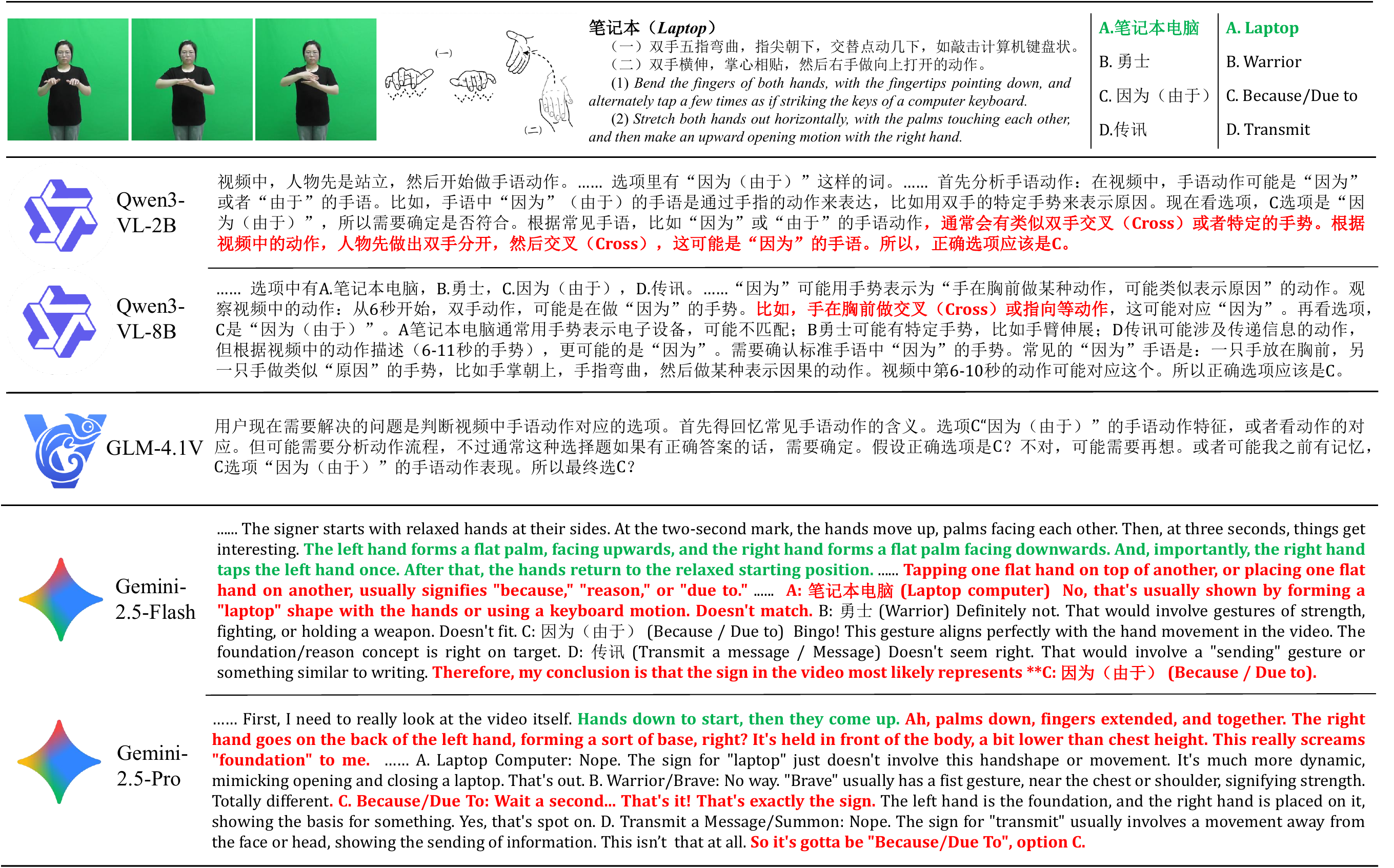}
\caption{Video-based sign language understanding failure for ``laptop''. Despite correctly recognizing the typing motion in a continuous sign language video, models fail to recover the intended meaning, indicating that temporal visual recognition alone is insufficient for reliable sign language understanding.}
\label{fig_case_study}
\end{figure*}


\subsection{Additional Analysis}

\paragraph{Test--Time Scaling.}\label{sec_cot}





To better contextualize empirical findings from the main results, we examine test-time scaling (reasoning) via CoT on MLLMs with explicit reasoning capabilities. We compare fast-thinking and slow-thinking settings across modalities and manual articulatory forms to assess when additional reasoning is beneficial for sign language understanding. 
After carefully analyzing these results, we draw several conclusions:
%

(1) \emph{CoT gains vary substantially across models.} 
As shown in Table~\ref{table_cot_results} and Figure~\ref{fig_slow_fast_1}, CoT gains vary substantially across models and subsets. Gemini-2.5-Flash consistently benefits most from slow thinking, achieving an average improvement of 6.45\%, whereas Qwen3-VL-Plus exhibits negative gains on several subsets, including textual descriptions, illustrative images, and sign language videos (with 10 frames per second). The repeated evaluations in Qwen3-VL-Plus yield consistent negative results, suggesting that reasoning does not universally improve performance and may be sensitive to model-specific inference behavior.

(2) \emph{CoT gains plateau for top-tier models.}
Figure~\ref{fig_slow_fast_2} shows that for top-tier models such as Gemini-2.5-Pro and GPT-5, increasing reasoning effort from low to high yields marginal or no gains, and can even degrade performance on sign language video inputs. This indicates a boundary effect of test-time scaling, where additional reasoning tokens no longer translate into improved understanding once a strong baseline is reached.

(3) \emph{Reasoning quality differs by modality.}
Figure~\ref{fig_kde} reveals that models exhibit large variance in reasoning length and modality preference. The reasoning tokens vary by nearly an order of magnitude across models under textual input. The Qwen3-VL and Gemini-2.5-Pro tend to allocate significantly more reasoning effort to text than to image or video inputs. Such modality-dependent behavior indicates incomplete multimodal alignment, which is especially problematic given the inherently visual-language nature of sign language. 

(4) \emph{Reasoning length correlates with task difficulty.} 
MLLMs tend to generate longer reasoning chains for incorrectly answered instances. The ratio between CoT length for incorrect versus correct predictions of GPT-5-M with text input reaches up to 2.89. This pattern is consistent across most models and modalities, suggesting that models spend more effort on harder cases, partially mirroring human problem-solving behavior. Please see Appendix~\ref{sec_app_cot} for detailed analysis.

These results indicate that CoT primarily serves as a diagnostic lens rather than a universal performance enhancer, revealing fundamental challenges in multimodal reasoning and sign language-specific understanding that persist in MLLMs.

\paragraph{Prompt Tokens \& Instruction Following.}
Beyond reasoning-oriented analyses, we further examine prompt token consumption and instruction-following robustness as complementary diagnostic factors for sign language understanding in MLLMs.
Prompt token usage varies substantially across input modalities and model families, with image and video inputs incurring orders-of-magnitude higher token consumption than text, potentially constraining effective context and reasoning budgets. 
In addition, instruction-following robustness differs markedly across models and modalities: while most large-scale models remain stable, several smaller-capacity models and certain MLLM families exhibit pronounced failures, particularly on sign language videos.
Notably, explicit reasoning mechanisms may also interact non-trivially with instruction adherence. 
Due to the space limitation, the detailed quantitative analyses are provided in Appendix~\ref{sec_app_prompt_tokens} and Appendix~\ref{sec_app_ins_follow}, respectively.
%

\subsection{Case Studies}
To provide intuitive insights into the sources of observed performance differences and further illustrate the limitations revealed by the CNSL-benchmark, we present a case study in Figure~\ref{fig_case_study}. A primary observation is the pronounced modality sensitivity. For the concept \emph{laptop}, the textual description explicitly mentions ``striking keys,'' allowing models to easily deduce the correct answer. However, under video inputs, even advanced models (e.g., Gemini-2.5-Pro, Qwen3-VL) fail to ground the visual motion correctly. Instead of recognizing the ``typing'' and ``opening'' gestures, they hallucinate unrelated motions (e.g., ``crossing'' gestures in Qwen3-VL-8B or ``foundation'' gestures in Gemini-2.5-Pro) and consistently misclassify the sign as ``Because'' (Option C). This contrast highlights that while MLLMs possess strong textual reasoning, their ability to parse complex temporal visual cues in sign language remains fragile.
Beyond modality effects, our analysis reveals challenges in implicit semantic association (e.g., ``smell'' vs ``air'') and emerging symbolic understanding (e.g., correctly mapping visual signs to Chinese characters), see cases in Appendix~\ref{sec_app_case_study}.


\section{Related Works}\label{sec_related_works}
\subsection{Sign Language Understanding}\label{sec_related_works_slu}

%
%

Over the past few decades, the sign language research community has primarily focused on sign language recognition (SLR), emphasizing the identification of gloss-level units\footnote{Glosses are spoken-language textual units that approximately capture the meaning of sign language.} in isolated words~\cite{jozeMSASLLargeScaleData2019, liWordlevelDeepSign2020} or in continuous sequences~\cite{kollerContinuousSignLanguage2015, huangVideobasedSignLanguage2018}. 
The increasing availability of large-scale sign language translation datasets~\cite{camgozNeuralSignLanguage2018, zhouImprovingSignLanguage2021, duarteHow2SignLargeScaleMultimodal2021,tanzerYouTubeSL25LargeScaleOpenDomain2024} has further driven progress in end-to-end sign language translation (SLT)~\cite{camgozMultichannelTransformersMultiarticulatory2020, chenSimpleMultiModalityTransfer2022, fuSignerDiversitydrivenData2024a, zhaoConditionalVariationalAutoencoder2024, zhangDynamicFeatureFusion2025, fuImprovingEndtoEndSign2025}.
In parallel, large language models (LLMs) have become general-NLP purpose backbones for a wide range of NLP tasks~\cite{brownLanguageModelsAre2020a,touvronLLaMAOpenEfficient2023a}, motivating recent efforts to incorporate them into sign language research for their strong language modeling and generation capabilities.
In particular, prior work has leveraged LLMs either as enhanced text decoders~\cite{wongSign2GPTLeveragingLarge2023, gongLLMsAreGood2024a, liu2024llavanext} or as semantic enhancement modules to improve SLT systems~\cite{guoBridgingSignSpoken2025a, kimLeveragingPowerMLLMs2025b, liuSCOPESignLanguage2025, jangLostTranslationFound2025a}. 

Existing work in this line is largely centered on task- or dataset-specific adaptation of LLMs within sign language pipelines. In contrast, systematic evaluation of models’ intrinsic sign language understanding, particularly for MLLMs operating directly on images and videos, has received comparatively less attention. 
Different from focusing on specific downstream tasks or datasets, we propose CNSL-bench, a comprehensive Chinese National Sign Language benchmark designed for evaluating MLLMs in sign language understanding.

\subsection{MLLM benchmarks}

Recent progress in multimodal large language models (MLLMs) has coincided with the establishment of standardized benchmarks designed to evaluate multimodal understanding. 
Early efforts primarily focused on image-based evaluation, including visual question answering, caption-based reasoning, and broader vision-language understanding benchmarks that assess perception, grounding, and semantic reasoning over static images~\cite{fuMMEComprehensiveEvaluation2025, yueMMMUMassiveMultiDiscipline2024, liSEEDBenchBenchmarkingMultimodal2024, chenM^3CoTNovelBenchmark2024}.
More recently, the community has extended benchmarking to video-centric settings, introducing datasets and evaluation protocols that emphasize temporal grounding, event understanding, long-context reasoning, and multimodal dialogue over videos~\cite{maazVideoChatGPTDetailedVideo2024, zhouMLVUBenchmarkingMultitask2025, fuVideoMMEFirstEverComprehensive2025}.

Despite their broad coverage, most existing MLLM benchmarks are designed for general-domain image and video understanding, where the visual content and semantics are dominated by everyday objects, scenes, actions, and events. Benchmarks that explicitly target sign language remain relatively scarce, despite the need for fine-grained modeling of hand articulation, motion trajectories, and linguistically grounded semantics. 
In contrast, CNSL-bench is constructed as a dedicated evaluation benchmark for sign language understanding, enabling systematic assessment of MLLMs under aligned textual descriptions, illustrative images, and sign language videos.

\section{Conclusion}

In this work, we introduce CNSL-bench, a multimodal benchmark centered on sign language for evaluating sign language understanding in MLLMs, with authoritative grounding in officially standardized sign language resources. 
Through extensive evaluation of a variety of open- and closed-source models, we systematically uncover persistent challenges in current MLLMs’ ability to comprehend sign language across modalities and manual articulatory forms. 
We anticipate that CNSL-bench will serve as both a diagnostic foundation and a reference resource for future research toward more robust, reliable, and human-aligned MLLMs.

\section*{Limitations}
CNSL-bench focuses on lexical-level canonical sign understanding and adopts a multiple-choice formulation to enable controlled, scalable, and reproducible evaluation across modalities.
While this design does not directly assess open-ended sign language generation, it is motivated by the observation that current MLLMs remain unreliable in interpreting free-form sign language.
This limitation highlights a fundamental gap between existing model capacities and the demands of open-ended sign interpretation, underscoring the necessity of establishing a diagnostic benchmark target at sign language understanding. 
In addition, CNSL-bench is centered on Chinese National Sign Language and emphasizes authoritative semantic grounding over linguistic breadth, and thus does not cover cross-linguistic, regional, or dialectal variation present in other sign languages. Extending evaluation to multilingual and multi-regional sign languages, as well as to more open-ended and compositional settings, remains an important direction for future work.

\section*{Ethical Considerations}

\paragraph{Data Access.}
CNSL-bench is constructed by aligning publicly available resources with recently released sign language video datasets and does not involve new data collection or additional human participants. The textual descriptions and illustrative images are sourced from officially published materials released by the Ministry of Education of the People’s Republic of China and are publicly accessible for educational and general communication purposes. 
The sign language videos are derived from an open-source dataset whose data collection and public release were approved by the Ethical Review Board of Leshan Normal University, with all participants providing informed consent for the use and publication of their identity information and recordings (Ethical Review Number: LSNU-KYLL2025-02-15)~\cite{jinLargeDatasetCovering2025}. 
\paragraph{Human Participant.}
Our benchmark involves the human assessment, and we invited a professional team consisting of one professor specializing in sign language linguistics and three sign-language students (including one hearing-impaired student). Each student has at least one year of classroom studying experience in sign language; their instructors include the invited professor and Deaf sign language teachers from a local special education institute. 
Participants are compensated with \$30 to complete each task (about two hours of work). Overall, one expert and three students are engaged to fulfill the human assessment tasks.

\section*{Acknowledgements}
We are grateful for the efforts and time of the reviewers and the committee. 
This work was supported in part by the National Natural Science Foundation of China under Grant 62476232, Grant 62076211, and in part by First Batch of Projects for the 2025 ``Intergovernmental International Science, Technology and Innovation Cooperation'' of the National Key Research and Development Program of China under Grant 2025YFE0121700.

\bibliography{custom}

\appendix




\section{CNSL-bench}\label{sec_app_cnsl}

\begin{figure*}[ht]
\centering
\includegraphics[width=1.0 \linewidth]{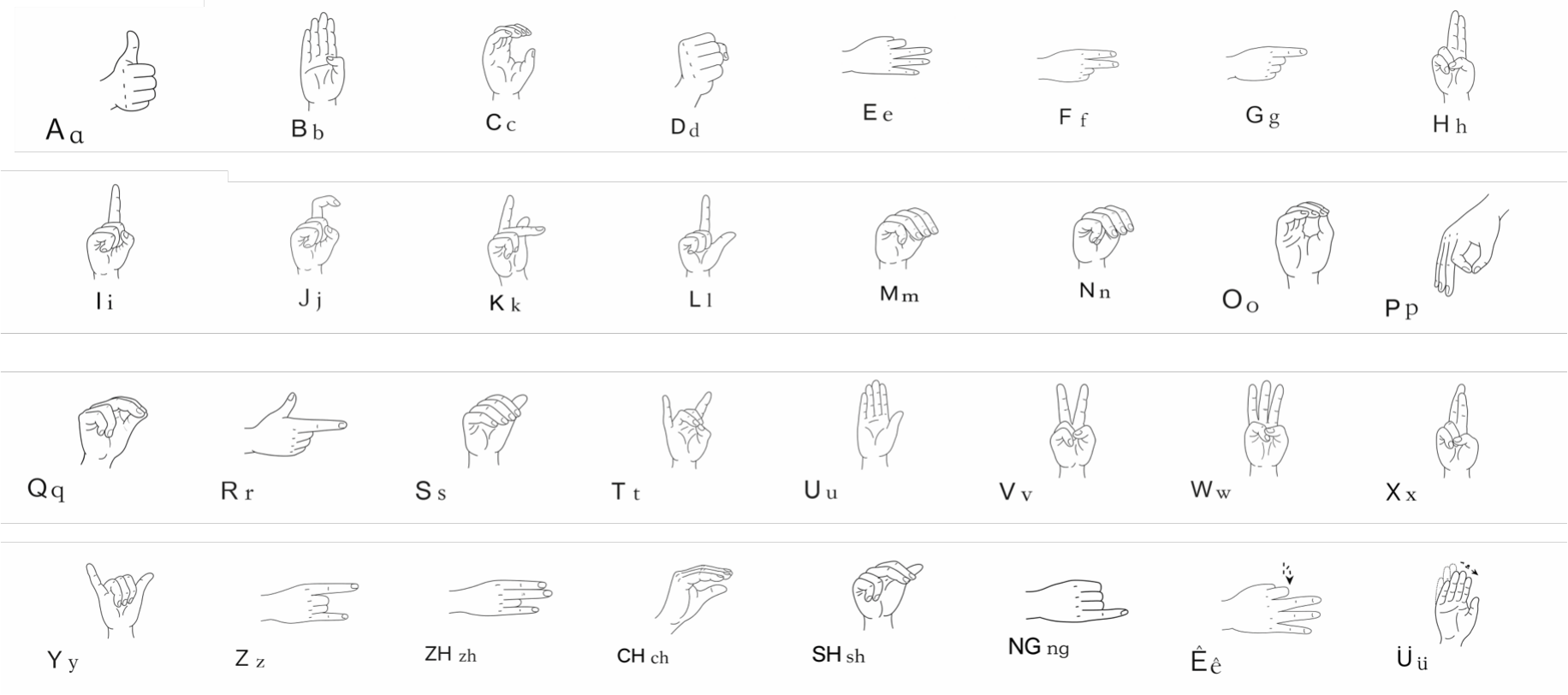}
\caption{All manual alphabets in the \textit{Chinese Manual Alphabet}, including 26 single-letter alphabets, 4 double-letter alphabets, and 2 alphabets with symbols.}
\label{fig_cnsl_alphabet}
\end{figure*}

\subsection{Dataset Alignment}\label{sec_app_cnsl_alignment}

This section details the data processing and alignment procedures underlying the construction of CNSL-bench. The textual descriptions and illustrative images are sourced from the National Common Sign Language Dictionary, which contains 8,214 commonly used sign entries and serves as an authoritative reference for standardized CSL~\cite{cnsl_common, cnsl_full}. During data preparation, we identify several forms of redundancy in the original dictionary. 
First, some entries share identical meanings and identical sign realizations, which are merged into a single sign entry. 
Second, certain lexical items correspond to multiple meanings, which are distinguished in the dictionary using auxiliary index markers. 
Third, some entries share the same lexical form and meaning but differ in their sign realizations. 
To ensure a consistent representation, we remove these auxiliary markers and retain all valid lexical variants associated with each sign realization, followed by sign-level processing. 
As a result, a set of 6,707 unique sign entries is obtained. 
In addition, the dictionary includes explanatory and instructional content intended for human readers, and we remove such descriptive text and preserve only the core lexical information. 
Moreover, the dictionary does not explicitly annotate linguistically manual articulatory forms such as air-writing, finger-spelling, or manual-alphabet, we manually identify and annotate these categories to support fine-grained analysis.

The sign language videos are sourced from the CNSL-DP dataset~\cite{jinLargeDatasetCovering2025}, which was collected under institutional ethical approval and provides synchronized video recordings for individual sign entries. For each sign, multiple video instances from different signers are available. To ensure both consistency and representativeness, we select one representative recording per sign entry for inclusion in the benchmark. 
The original videos are recorded at a resolution of 1920 $\times$ 1080 and 50 frames per second, with the signer centered in the frame. We uniformly downsample the videos to 24 frames per second and apply center cropping followed by resizing to 512 $\times$ 512 to standardize visual inputs. 
In cases where multiple synonymous lexical entries correspond to the same sign realization, the original CNSL-DP dataset retains only a single representative form. To construct a lexically complete benchmark aligned with the official dictionary, we explicitly recover the omitted synonymous entries and re-associate them with the corresponding video instances during dataset alignment, thereby restoring the full set of lexical variants for each sign. As a result, CNSL-bench establishes a unified mapping among textual descriptions, illustrative images, and sign language videos, thereby constructing a coherent and well-aligned benchmark for evaluating sign language understanding.

\subsection{The China Manual Alphabet}\label{sec_app_cnsl_alphabet}

Figure~\ref{fig_cnsl_alphabet} shows all manual alphabets in the Chinese Manual Alphabet, including 26 single-letter manual alphabets, 4 double-letter manual alphabets, and 2 manual alphabets with symbols.

\begin{figure*}[t]
\centering
\includegraphics[width=1.0 \linewidth]{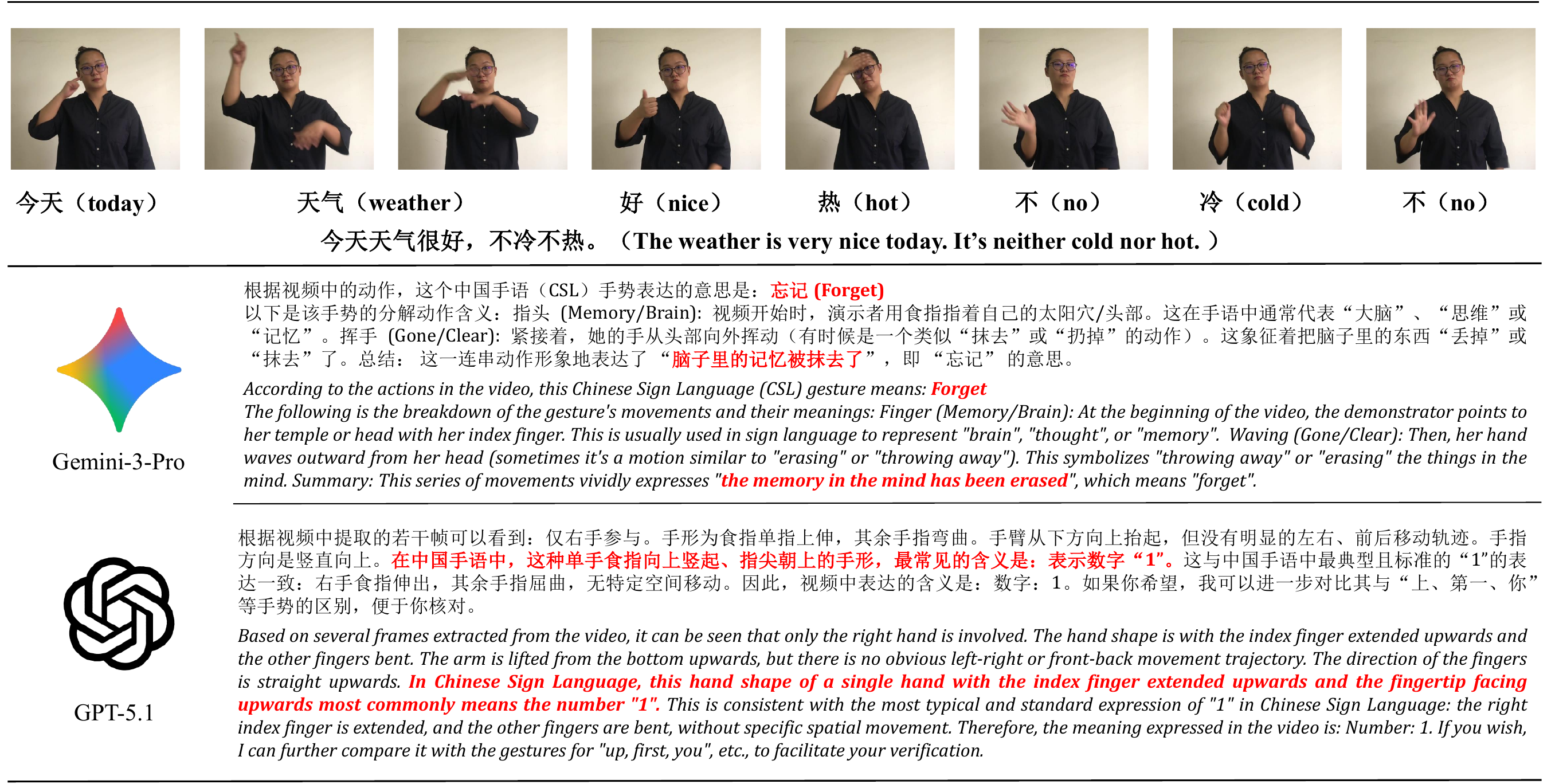}
\caption{
An example of open-ended sign language understanding from the CSL-Daily dataset~\cite{zhouImprovingSignLanguage2021}. The input video expresses a simple sentence (``The weather is very nice today, neither cold nor hot''), yet flagship models (e.g., Gemini-3-Pro and GPT-5.1) fail to generate the correct meaning. Note that the realization of concepts such as ``weather'' in this example differs from the canonical forms used in CNSL-bench, reflecting natural variation in sign expression, while remaining readily interpretable to human signers.
}
\label{fig_case_open_ended}
\end{figure*}

\subsection{Detailed Task Definition}\label{sec_app_task}

This section provides additional description supporting the task formulation and option construction strategies adopted in CNSL-bench. We first examine the feasibility of open-ended sign language understanding and find that current state-of-the-art MLLMs remain highly unreliable in this setting. As illustrated in Figure~\ref{fig_case_open_ended}, we draw a case from CSL-Daily~\cite{zhouImprovingSignLanguage2021}, a Chinese sign language dataset targeted at sign language recognition and translation. The results show that even for a short video expressing a simple and common sentence (e.g., ``The weather is very nice today, neither cold nor hot''), flagship models such as Gemini-3-Pro and GPT-5.1 fail to recover the intended meaning, instead misinterpreting isolated handshapes or hallucinating unrelated lexical concepts. The observed failures suggest that existing models struggle to robustly integrate temporal dynamics, sequential articulation, and lexical composition in free-form generation, leading to unstable and difficult-to-interpret outputs. 

\begin{table}[ht]
\resizebox{1.0 \linewidth}{!} {
\begin{tabular}{@{}lcccc@{}}
\toprule
Distractor & Text & Image & Video$^{'2}$ & Video$^{'10}$ \\ 
\midrule
Random & 72.41 & 38.83 & 32.06 & 33.95 \\ 
Semantic-Based & 64.33 & 36.95 & 32.60 & 32.47 \\ 
\bottomrule
\end{tabular}
}
\caption{An analysis of distractor on Qwen-VL-Plus.}
\label{table_distractor}
\end{table}

In addition, we further analyze different option construction strategies for the multiple-choice formulation. As shown in Table~\ref{table_distractor}, semantics-based distractors yield slightly lower absolute accuracy than random sampling across modalities on a representative model (i.e., Qwen-VL-Plus), yet they lead to qualitatively consistent conclusions about model performance. This indicates that the overall ranking and modality-dependent trends are stable under alternative distractor designs, and that benchmark outcomes are not driven by overly obvious negatives. In other words, random sampling already provides sufficiently challenging and reliable evaluation signals for CNSL-bench. Since semantics-based distractors introduce additional engineering overhead and potential sensitivity to similarity heuristics without changing the main findings, we adopt random option sampling to streamline the benchmark design and support robust, reproducible evaluation.

\subsection{Detailed Dataset Statistics}\label{sec_app_statistics}

The detailed statistics of CNSL-bench are illustrated in Table~\ref{table_statistics}, and the frame count is calculated at a rate of 24 frames per second.

\begin{table}[h]
\centering
\begin{tabular}{@{}lrr@{}}
\toprule
Subset & \multicolumn{1}{c}{\#Sign Entry} & \multicolumn{1}{c}{\#Frames} \\ \midrule
Air-Writing & 407 & 99.8 \\
Finger-Spelling & 77 & 109.2 \\
Manual-Alphabet  & 592 & 98.0 \\ \midrule
\textit{w/} $1$ gesture & 2,977 & 89.2 \\
\textit{w/} $2$ gestures & 3,287 & 100.2 \\
\textit{w/} $3$ gestures & 369 & 120.9 \\
\textit{w/} $4$ gestures & 62 & 136.4 \\
\textit{w/} $5$ gestures & 8 & 148.4 \\
\textit{w/} $6$ gestures & 2 & 161.0 \\
\textit{w/} $7$ gestures & 2 & 201.5 \\ \midrule
All & 6707 & 96.89 \\ \bottomrule
\end{tabular}
\caption{The detailed statistics of CNSL-bench.}
\label{table_statistics}
\end{table}

\section{Experiments}
\subsection{Experimental Settings}\label{sec_app_settings}

\paragraph{MLLMs Participants.} A total of 21 MLLMs (13 Open-source and 8 closed-source MLLMs) are included for validation, which includes 
a) 3 open\&closed-source Image MLLMs: LLaVA-NeXT (Mistral-7B)~\cite{liu2024llavanext}, Qwen-VL-Plus/Max~\cite{baiQwenVLVersatileVisionLanguage2023}, 
b) 12 open-source MLLMs: Qwen2-VL-2B/7B~\cite{wangQwen2VLEnhancingVisionLanguage2024}, Qwen2.5-VL-3B/7B~\cite{baiQwen25VLTechnicalReport2025}, Intern3.5-VL-2B/8B~\cite{wangInternVL35AdvancingOpenSource2025}, Qwen3-VL-2B/8B-Instruct, Qwen3-VL-2B/8B-Thinking~\cite{baiQwen3VLTechnicalReport2025a}, LLaVA-NeXT-Video-7B~\cite{liu2024llavanext}, and GLM-4.1V-9B-Thinking~\cite{teamGLM45VGLM41VThinkingVersatile2025}, 
and c) 6 close-source MLLMs: Qwen3-VL-Plus~\cite{baiQwen3VLTechnicalReport2025a}, Gemini-2.5-Flash, Gemini-2.5-Pro~\cite{comaniciGemini25Pushing2025}, GPT-4o-mini, GPT-4o, and GPT-5~\cite{openaiGPT4TechnicalReport2024}.

\paragraph{Evaluation Details.} For open source MLLMs, we primarily use the Hugging Face transformers library\footnote{\url{https://hugging-face.cn/docs/transformers}} for model inference. 
To accelerate decoding for \emph{thinking} models under the slow-thinking setting, we adopt vLLM\footnote{\url{https://github.com/vllm-project/vllm}} as the inference backend, and set the maximum generation length to 8,192 tokens per response.
For closed-source MLLMs, any samples that fail due to API errors, timeouts, or malformed generations are excluded from scoring, so that the reported metric scores are computed only over valid outputs.
As for video inputs, we evaluate two frame sampling rates: 2 fps (the default in many video-understanding benchmarks) and a denser 10 fps setting, which better captures the high-speed and fine-grained spatiotemporal motions characteristic of sign language.

\begin{table*}[t]
\resizebox{1.0 \linewidth}{!} {
\begin{tabular}{@{}lrrrcrrrcrrrcrrrc@{}}
\toprule
\multirow{2}{*}{Model} & \multicolumn{4}{c}{Text} & \multicolumn{4}{c}{Image} & \multicolumn{4}{c}{Video$^{'2~fps}$} & \multicolumn{4}{c}{Video$^{'10~fps}$} \\ 
\cmidrule(lr){2-5} \cmidrule(lr){6-9} \cmidrule(lr){10-13} \cmidrule(l){14-17}
 & Correct & Fault & All & Ratio & Correct & Fault & All & Ratio & Correct & Fault & All & Ratio & Correct & Fault & All & Ratio \\ \midrule
Qwen3-VL-8B & 1,688 & 2,905 & 2,046 & 1.72 & 255 & 303 & 285 & 1.19 & 390 & 414 & 407 & 1.06 & 252 & 275 & 267 & 1.09 \\
GLM-4.1V-9B & 179 & 304 & 218 & 1.70 & 278 & 379 & 339 & 1.36 & 335 & 400 & 382 & 1.19 & 318 & 397 & 374 & 1.25 \\
Qwen3-VL-Plus & 1,212 & 2,124 & 1,429 & 1.75 & 360 & 431 & 401 & 1.20 & 1,264 & 1,411 & 1,359 & 1.12 & 322 & 317 & 319 & 0.98 \\
Gemini-2.5-Flash & 818 & 1,755 & 1,006 & 2.15 & 1,090 & 1,827 & 1,447 & 1.67 & 719 & 938 & 845 & 1.30 & 720 & 934 & 843 & 1.30 \\
Gemini-2.5-Pro (L) & 347 & 349 & 347 & 1.00 & 132 & 135 & 133 & 1.02 & 76 & 73 & 74 & 0.95 & 82 & 84 & 83 & 1.02 \\
Gemini-2.5-Pro (M) & 1,133 & 1,755 & 1,228 & 1.55 & 941 & 1,281 & 1,073 & 1.36 & 700 & 788 & 746 & 1.13 & 708 & 803 & 757 & 1.14 \\
Gemini-2.5-Pro (H) & 365 & 532 & 390 & 1.46 & 915 & 1,177 & 1,015 & 1.29 & 675 & 757 & 718 & 1.12 & 699 & 797 & 749 & 1.14 \\
GPT-5 (L) & 248 & 637 & 291 & 2.57 & 339 & 494 & 391 & 1.45 & 325 & 395 & 359 & 1.22 & 353 & 442 & 394 & 1.25 \\
GPT-5 (M) & 635 & 1,837 & 759 & 2.89 & 1,017 & 1,614 & 1,214 & 1.59 & 1,093 & 1,420 & 1,245 & 1.30 & 1,312 & 1,701 & 1,480 & 1.30 \\
GPT-5 (H) & 1,392 & 3,737 & 1,628 & 2.68 & 2,248 & 3,211 & 2,553 & 1.43 & 2,305 & 2,777 & 2,527 & 1.20 & 2,597 & 3,150 & 2,852 & 1.21 \\ 
\bottomrule
\end{tabular}
}
\caption{Detailed reasoning tokens across models and modalities. L, M, H: low, medium, and high reasoning effort on the process of thinking before generating an answer.}
\label{table_cot_token}
\centering
\end{table*}

To mitigate input length constraints, when a dense sampling rate (e.g., FPS=10) would exceed a model’s maximum context window, we adaptively resample the video and include as many frames as possible \emph{without} surpassing the input length limit (e.g., for LLaVA-NeXT-Video, we reduce the sampling rate but maximize the number of frames allowed within its context budget). 
For the GPT-series models, we uniformly cap the visual input to at most 50 frames to comply with the API restriction.
Unless stated otherwise, all hyperparameters follow the official recommendations for each model, as summarized in Table~\ref{table_hyperparameters}.
Accuracy is computed by an exact match between the model prediction and the ground-truth answer. For thinking models, we manually extract the final answer from the generated response to avoid conflating intermediate reasoning with the predicted label. 

\begin{table}[ht]
\resizebox{1.0 \linewidth}{!} {
\begin{tabular}{@{}lcccccc@{}}
\toprule
\multirow{2}{*}{Model} & \multicolumn{3}{c}{Text} & \multicolumn{3}{c}{V-L} \\ \cmidrule(r){2-4} \cmidrule(){5-7} 
 & top\_p & top\_k & T & top\_p & top\_k & T \\ 
\midrule
LLaVA-NeXT-7B & 0.95 & 50 & 1.0 & 0.95 & 50 & 1.0 \\
LLaVA-NeXT-Video-7B & 0.95 & 50 & 1.0 & 0.95 & 50 & 1.0 \\
Qwen2/2.5-VL-Instruct & 0.95 & 50 & 1.0 & 0.95 & 50 & 1.0 \\
Intern-VL-3.5 & 0.95 & 50 & 1.0 & 0.95 & 50 & 1.0 \\
GLM-4.1V-9B & 0.95 & 50 & 1.0 & 0.95 & 50 & 1.0 \\
Qwen3-VL-Instruct & 1.00 & 40 & 1.0 & 0.80 & 20 & 0.7 \\
Qwen3-VL & 0.95 & 20 & 1.0 & 0.95 & 20 & 1.0 \\
Gemini-2.5-Series & 0.95 & - & 1.0 & 0.95 & - & 1.0 \\
GPT-Series & 1.0 & - & 1.0 & 1.0 & - & 1.0 \\ 
\bottomrule
\end{tabular}
}
\caption{Detailed hyperparameters. T means temperature. V-L denotes multimodal input settings.}
\label{table_hyperparameters}
\end{table}


\paragraph{Human Assessment.} 
%
Deaf community involvement is essential for developing sign language understanding systems~\cite{yinIncludingSignedLanguages2021, atwellStudyingMitigatingBiases2024}. To establish a human reference for CNSL-bench, we invited a professional team consisting of one professor specializing in sign language linguistics and three sign-language students (including one hearing-impaired student). Each student has at least one year of classroom studying experience in sign language; their instructors include the invited professor and Deaf sign language teachers from a local special education institute. 
To make the evaluation feasible while preserving articulation diversity, we constructed a 1,500-entry subset from the 6,707 sign entries. Specifically, we retained all entries involving air-writing, finger-spelling, and manual-alphabet articulations, yielding 1,018 entries, and then randomly sampled an additional 482 entries from the remaining gesture-only entries. For each entry, we generated three multiple-choice questions corresponding to the aligned textual description, illustrative image, and sign language video, resulting in 4,500 questions in total. Each evaluator completed all questions, and we report the average accuracy over the three student evaluators as the human performance.

\begin{table*}[ht]
\resizebox{1.0 \linewidth}{!} {
\begin{tabular}{lccccccccccccccccc}
\toprule
\multirow{2}{*}{Model} & \multicolumn{4}{c}{Text} & \multicolumn{4}{c}{Image} & \multicolumn{4}{c}{Video$^{'2~fps}$} & \multicolumn{4}{c}{Video$^{'10~fps}$} \\ 
\cmidrule(r){2-5} \cmidrule(lr){6-9} \cmidrule(lr){10-13} \cmidrule(l){14-17}
& AW & FS & MA & All & AW & FS & MA & All & AW & FS & MA & All & AW & FS & MA & All \\ 
\cmidrule(){1-17}
\multicolumn{17}{c}{\cellcolor[HTML]{EFEFEF}\textit{Open\& Close -source Image MLLMs}} & \multirow{25}{*}{%
  \begin{tikzpicture}
    \fill[
        bottom color=red!80,
        middle color=yellow,
        top color=green!45!yellow,
    ] (-0.5,0.7) rectangle (0, 13.5);
    \foreach \y/\label in {
        0.9/{0\%}, 
        7/{50\%},
        13.3/{100\%}
    } {\draw (0,\y) node[right]{\label};}
  \end{tikzpicture}} \\ 
\cmidrule(){1-17}
\cellcolor{yellow!45!orange}{LLaVA-NeXT-7B}
& \accHeat{1.2} & \accHeat{6.5} & \accHeat{2.5} & \accHeat{4.9}
& \accHeat{61.2} & \accHeat{54.6} & \accHeat{62.3} & \accHeat{62.8}
& \accHeat{68.8} & \accHeat{71.4} & \accHeat{73.7} & \accHeat{71.2}
& \accHeat{-} & \accHeat{-} & \accHeat{-} & \accHeat{-} 
\\
\cellcolor{green!40}{Qwen-VL-Plus}
& \accHeat{100.0} & \accHeat{100.0} & \accHeat{100.0} & \accHeat{100.0}
& \accHeat{100.0} & \accHeat{100.0} & \accHeat{100.0} & \accHeat{100.0}
& \accHeat{100.0} & \accHeat{100.0} & \accHeat{100.0} & \accHeat{99.9}
& \accHeat{99.8} & \accHeat{100.0} & \accHeat{100.0} & \accHeat{99.9} \\
\cellcolor{green!45}{Qwen-VL-Max}
& \accHeat{100.0} & \accHeat{100.0} & \accHeat{100.0} & \accHeat{100.0}
& \accHeat{100.0} & \accHeat{100.0} & \accHeat{100.0} & \accHeat{100.0}
& \accHeat{100.0} & \accHeat{100.0} & \accHeat{100.0} & \accHeat{100.0}
& \accHeat{100.0} & \accHeat{100.0} & \accHeat{100.0} & \accHeat{100.0} \\ 
\cmidrule(){1-17}
\multicolumn{17}{c}{\cellcolor[HTML]{EFEFEF}\textit{Open-Source MLLMs}} \\ 
\cmidrule(){1-17}
\cellcolor{green!30!yellow}{Qwen2-VL-2B}
& \accHeat{85.5} & \accHeat{83.1} & \accHeat{85.0} & \accHeat{83.4}
& \accHeat{98.5} & \accHeat{97.4} & \accHeat{99.0} & \accHeat{98.3}
& \accHeat{100.0} & \accHeat{100.0} & \accHeat{100.0} & \accHeat{99.9}
& \accHeat{97.1} & \accHeat{90.9} & \accHeat{98.3} & \accHeat{97.7} \\
\cellcolor{green!45}{Qwen2.5-VL-3B}
& \accHeat{100.0} & \accHeat{100.0} & \accHeat{100.0} & \accHeat{100.0}
& \accHeat{100.0} & \accHeat{100.0} & \accHeat{100.0} & \accHeat{100.0}
& \accHeat{100.0} & \accHeat{100.0} & \accHeat{100.0} & \accHeat{100.0}
& \accHeat{100.0} & \accHeat{100.0} & \accHeat{100.0} & \accHeat{100.0} \\
\cellcolor{yellow!60!orange}{Intern-VL-3.5-2B}
& \accHeat{99.8} & \accHeat{100.0} & \accHeat{99.7} & \accHeat{99.4}
& \accHeat{99.3} & \accHeat{97.4} & \accHeat{97.8} & \accHeat{97.5}
& \accHeat{15.2} & \accHeat{15.6} & \accHeat{16.4} & \accHeat{15.3}
& \accHeat{17.4} & \accHeat{14.3} & \accHeat{16.4} & \accHeat{15.4} \\

\cellcolor{green!45}{Qwen3-VL-2B}
& \accHeat{99.8} & \accHeat{100.0} & \accHeat{100.0} & \accHeat{99.9}
& \accHeat{100.0} & \accHeat{100.0} & \accHeat{100.0} & \accHeat{100.0}
& \accHeat{100.0} & \accHeat{100.0} & \accHeat{100.0} & \accHeat{100.0}
& \accHeat{100.0} & \accHeat{100.0} & \accHeat{100.0} & \accHeat{100.0} \\ 
\cellcolor{yellow!45!orange}{LLaVA-NeXT-Video-7B}
& \accHeat{3.2} & \accHeat{3.9} & \accHeat{3.7} & \accHeat{4.8}
& \accHeat{45.2} & \accHeat{44.2} & \accHeat{47.1} & \accHeat{49.8}
& \accHeat{63.6} & \accHeat{58.4} & \accHeat{63.7} & \accHeat{64.0}
& \accHeat{60.0} & \accHeat{58.4} & \accHeat{63.3} & \accHeat{62.6} \\
\cellcolor{green!45}{Qwen2-VL-7B}
& \accHeat{100.0} & \accHeat{100.0} & \accHeat{100.0} & \accHeat{100.0}
& \accHeat{100.0} & \accHeat{100.0} & \accHeat{100.0} & \accHeat{100.0}
& \accHeat{100.0} & \accHeat{100.0} & \accHeat{100.0} & \accHeat{100.0}
& \accHeat{100.0} & \accHeat{100.0} & \accHeat{100.0} & \accHeat{100.0} \\
\cellcolor{green!45}{Qwen2.5-VL-7B}
& \accHeat{100.0} & \accHeat{100.0} & \accHeat{100.0} & \accHeat{100.0}
& \accHeat{100.0} & \accHeat{100.0} & \accHeat{100.0} & \accHeat{100.0}
& \accHeat{100.0} & \accHeat{100.0} & \accHeat{100.0} & \accHeat{100.0}
& \accHeat{100.0} & \accHeat{100.0} & \accHeat{100.0} & \accHeat{100.0} \\
\cellcolor{green!45}{GLM-4.1V-9B \ding{45}}
& \accHeat{99.5} & \accHeat{100.0} & \accHeat{99.7} & \accHeat{99.8}
& \accHeat{98.8} & \accHeat{98.7} & \accHeat{99.3} & \accHeat{99.1}
& \accHeat{99.5} & \accHeat{100.0} & \accHeat{99.2} & \accHeat{99.5}
& \accHeat{99.3} & \accHeat{98.7} & \accHeat{99.2} & \accHeat{99.4} \\ 
\cellcolor{green!45}{Intern-VL-3.5-8B}
& \accHeat{100.0} & \accHeat{100.0} & \accHeat{100.0} & \accHeat{100.0}
& \accHeat{100.0} & \accHeat{100.0} & \accHeat{100.0} & \accHeat{100.0}
& \accHeat{100.0} & \accHeat{100.0} & \accHeat{100.0} & \accHeat{100.0}
& \accHeat{100.0} & \accHeat{100.0} & \accHeat{100.0} & \accHeat{100.0} \\
\cellcolor{green!45}{Qwen3-VL-8B-Instruct}
& \accHeat{100.0} & \accHeat{100.0} & \accHeat{100.0} & \accHeat{100.0}
& \accHeat{100.0} & \accHeat{100.0} & \accHeat{100.0} & \accHeat{100.0}
& \accHeat{100.0} & \accHeat{100.0} & \accHeat{100.0} & \accHeat{100.0}
& \accHeat{100.0} & \accHeat{100.0} & \accHeat{100.0} & \accHeat{100.0} \\
\cellcolor{green!45}{Qwen3-VL-8B}
& \accHeat{100.0} & \accHeat{100.0} & \accHeat{100.0} & \accHeat{100.0}
& \accHeat{100.0} & \accHeat{100.0} & \accHeat{100.0} & \accHeat{100.0}
& \accHeat{100.0} & \accHeat{100.0} & \accHeat{100.0} & \accHeat{100.0}
& \accHeat{100.0} & \accHeat{100.0} & \accHeat{100.0} & \accHeat{100.0} \\
\cellcolor{green!45}{Qwen3-VL-8B \ding{45}}
& \accHeat{99.0} & \accHeat{96.1} & \accHeat{94.9} & \accHeat{99.1}
& \accHeat{100.0} & \accHeat{100.0} & \accHeat{100.0} & \accHeat{100.0}
& \accHeat{100.0} & \accHeat{100.0} & \accHeat{99.8} & \accHeat{100.0}
& \accHeat{100.0} & \accHeat{100.0} & \accHeat{100.0} & \accHeat{100.0} \\
\cmidrule(){1-17}
\multicolumn{17}{c}{\cellcolor[HTML]{EFEFEF}\textit{Closed-Source MLLMs}} \\ 
\cmidrule(){1-17}
\cellcolor{green!45}{Qwen3-VL-Plus \ding{45}}
& \accHeat{100.0} & \accHeat{100.0} & \accHeat{100.0} & \accHeat{99.9}
& \accHeat{100.0} & \accHeat{100.0} & \accHeat{100.0} & \accHeat{100.0}
& \accHeat{100.0} & \accHeat{100.0} & \accHeat{100.0} & \accHeat{100.0}
& \accHeat{100.0} & \accHeat{100.0} & \accHeat{100.0} & \accHeat{100.0} \\
\cellcolor{green!45}{Gemini-2.5-Flash}
& \accHeat{100.0} & \accHeat{100.0} & \accHeat{100.0} & \accHeat{100.0}
& \accHeat{100.0} & \accHeat{100.0} & \accHeat{100.0} & \accHeat{100.0}
& \accHeat{100.0} & \accHeat{100.0} & \accHeat{100.0} & \accHeat{100.0}
& \accHeat{100.0} & \accHeat{100.0} & \accHeat{100.0} & \accHeat{100.0} \\
\cellcolor{green!40}{Gemini-2.5-Flash \ding{45}}
& \accHeat{99.8} & \accHeat{100.0} & \accHeat{99.3} & \accHeat{99.8}
& \accHeat{92.9} & \accHeat{92.2} & \accHeat{92.9} & \accHeat{96.3}
& \accHeat{98.3} & \accHeat{96.1} & \accHeat{97.5} & \accHeat{98.4}
& \accHeat{97.5} & \accHeat{98.7} & \accHeat{99.0} & \accHeat{98.4} \\
\cellcolor{green!45}{Gemini-2.5-Pro \ding{45}}
& \accHeat{100.0} & \accHeat{100.0} & \accHeat{100.0} & \accHeat{100.0}
& \accHeat{100.0} & \accHeat{100.0} & \accHeat{100.0} & \accHeat{100.0}
& \accHeat{100.0} & \accHeat{100.0} & \accHeat{100.0} & \accHeat{100.0}
& \accHeat{100.0} & \accHeat{100.0} & \accHeat{100.0} & \accHeat{100.0} \\
\cellcolor{green!45}{GPT-4o-mini}
& \accHeat{100.0} & \accHeat{100.0} & \accHeat{100.0} & \accHeat{100.0}
& \accHeat{100.0} & \accHeat{100.0} & \accHeat{100.0} & \accHeat{100.0}
& \accHeat{100.0} & \accHeat{100.0} & \accHeat{100.0} & \accHeat{100.0}
& \accHeat{-} & \accHeat{-} & \accHeat{-} & \accHeat{-} \\
\cellcolor{green!45}{GPT-4o}
& \accHeat{99.8} & \accHeat{98.7} & \accHeat{99.8} & \accHeat{99.8}
& \accHeat{98.8} & \accHeat{100.0} & \accHeat{99.2} & \accHeat{98.9}
& \accHeat{99.5} & \accHeat{100.0} & \accHeat{99.3} & \accHeat{99.2}
& \accHeat{99.0} & \accHeat{98.7} & \accHeat{98.3} & \accHeat{98.6} \\
\cellcolor{green!45}{GPT-5 \ding{45}} 
& \accHeat{100.0} & \accHeat{100.0} & \accHeat{100.0} & \accHeat{100.0}
& \accHeat{100.0} & \accHeat{100.0} & \accHeat{100.0} & \accHeat{100.0}
& \accHeat{100.0} & \accHeat{100.0} & \accHeat{100.0} & \accHeat{100.0}
& \accHeat{100.0} & \accHeat{100.0} & \accHeat{100.0} & \accHeat{100.0} \\ 
\bottomrule
\end{tabular}
}
\caption{Instruction Following. \ding{45} denotes inference with slow thinking.}
\label{table_instruction_following}
\centering
\end{table*}

\begin{table}[ht!]
\resizebox{1.0 \linewidth}{!} {
\centering
\begin{tabular}{@{}lrrrr@{}}
\toprule
Model & Text & Image & Video$^{'2}$ & Video$^{'10}$ \\ \midrule
\multicolumn{5}{c}{\cellcolor[HTML]{EFEFEF}\textit{Open-Source MLLMs}} \\ \midrule
LLaVA-NeXT-7B & 272 & 2,453 & 8,913 & - \\ \cmidrule(){2-5} 
LLaVA-NeXT-Video \ding{34} & 307 & 2,480 & 1,306 & 3,698 \\
Qwen2/2.5-VL & 169 & 742 & 1,413 & 6,669 \\
Intern-VL-3.5 & 195 & 415 & 2,102 & 10,659 \\
Qwen3-VL & 158 & 590 & 1163 & 5,445 \\
GLM-4.1V-9B\ding{45} & 153 & 726 & 1,409 & 6,713 \\ \midrule
\multicolumn{5}{c}{\cellcolor[HTML]{EFEFEF}\textit{Closed-Source MLLMs}} \\ \midrule
Qwen-VL-Plus & 218 & 1,519 & 3,552 & 15,586 \\
Qwen-VL-Max & 218 & 1,621 & 3,673 & 16,892 \\
Qwen3-VL-Plus & 206 & 1,350 & 3,375 & 20,842 \\
Qwen3-VL-Plus\ding{45} & 210 & 1,396 & 3,228 & 14,334 \\ 
\cmidrule(){2-5} 
Gemini-2.5-Flash & 225 & 4,388 & 3,124 & 3,343 \\
Gemini-2.5-Flash\ding{45} & 206 & 2,025 & 3,043 & 3,022 \\
Gemini-2.5-Pro\ding{45} & 198 & 2,576 & 2,501 & 2,451 \\
GPT-4o-mini & 270 & 71,274 & 230,971 & - \\
GPT-4o & 264 & 1,993 & 5,507 & 8,383 \\
GPT-5\ding{45} & 202 & 1,161 & 3,219 & 8,382 \\ 
\bottomrule
\end{tabular}
}
\caption{Average prompt token consumption across different input modalities.
\ding{45} denotes inference with slow thinking. \ding{34} indicates that videos are sampled at a maximum of 6 FPS due to the context window limitation.}
\label{table_prompt_tokens}
\end{table}

\section{Additional Analysis}

\subsection{Detailed Reasoning Tokens}\label{sec_app_cot}

Table~\ref{table_cot_token} reports the average reasoning tokens generated by each model, stratified by correctness and modality. Across all evaluated MLLMs, incorrect predictions are consistently associated with substantially longer reasoning traces than correct ones, reinforcing the observation that models tend to ``think longer'' when facing harder or ambiguous inputs, partially mirroring human problem-solving behavior. This effect is particularly pronounced for stronger models. For instance, GPT-5 (M) exhibits a ratio of 2.89 between incorrect and correct cases under text input, indicating that failed attempts often trigger nearly three times as many reasoning tokens. Similar trends are observed in Gemini-2.5-Flash, whose ratios exceed 2.0 in the text modality. 

This phenomenon generalizes to image and video inputs, albeit with a noticeably attenuated magnitude. 
In image settings, the ratios between incorrect and correct reasoning length typically fall within 1.2--1.7, while with video input, the ratios further decrease to around 1.0--1.3.
This compression suggests that the tendency to engage in longer reasoning on more difficult cases, which is clearly observed in the text-only setting, becomes less pronounced once multimodal perception is introduced. 
Rather than reflecting increased confidence, the reduced gap more plausibly indicates that multimodal perception and alignment imperfections constrain the model’s ability to adaptively allocate reasoning effort: when visual evidence is noisy, underspecified, or imperfectly aligned with the language space, the model may fail to trigger longer, exploratory reasoning even on genuinely difficult instances.
A further supporting signal is that increasing the temporal resolution of video inputs does not yield a systematic restoration of the gap. The ratios observed at 2 FPS and 10 FPS remain highly similar across models, including GPT-5 and Gemini-2.5 variants, despite the substantially increased number of frames. This suggests that the attenuation is not primarily driven by insufficient temporal evidence, but rather by broader limitations in multimodal understanding, such as imperfect robustness in visual feature extraction, temporal integration, or cross-modal grounding, which prevent additional frames from translating into more accurately calibrated reasoning effort.

Conclusively, the results point to a modality-dependent divergence in test-time behavior specific to sign language understanding. Although MLLMs display difficulty-sensitive processing in text-based settings, this characteristic is notably attenuated for visual inputs. Such attenuation suggests that current models may struggle to effectively utilize visual linguistic cues, indicating that limitations in multimodal perception and alignment contribute to the reduced adaptability observed in sign language understanding tasks.


\subsection{Detailed Prompt Tokens}\label{sec_app_prompt_tokens}

As shown in Table~\ref{table_prompt_tokens}, the consumption of prompt tokens varies substantially across both input modalities and models.
For a fixed model, image and video inputs consistently generate orders of magnitude more prompt tokens than text, resulting in an extreme length gap in multimodal processing.
This disparity in multimodal processing may partially account for the observed performance discrepancies across modalities.
Beyond modality effects, the token usage also differs markedly under identical text inputs, which can be largely attributed to heterogeneous tokenization and visual encoding strategies (e.g., LLaVa-NeXT vs. Qwen). Such tokenizer-induced prompt length variations may further affect effective context allocation and reasoning budget, introducing an additional source of performance variability in cross-model comparisons. 
Finally, closed-source models exhibit distinct budget characteristics shaped by their pricing-oriented design choices. Although GPT-4o-mini offers a lower per-token cost, its substantially higher token consumption for multimodal inputs results in significantly increased overall usage, leading us to exclude it from further evaluation due to prohibitive cost considerations.

\subsection{Instruction Following}\label{sec_app_ins_follow}


%

As shown in Table~\ref{table_instruction_following}, instruction adherence varies substantially across models and input modalities. While most large-scale open-source and proprietary MLLMs achieve near-perfect instruction-following accuracy across settings, several smaller-capacity models and certain MLLM families exhibit pronounced failures. 
Specifically, InternVL-3.5-2B maintains high accuracy on text and image inputs but collapses to around 15\% accuracy on sign language videos, indicating severe difficulty in jointly satisfying visual, temporal, and task-level constraints. In contrast, an opposite pattern is observed in Qwen2-VL-2B and models from the LLaVA-Next family, where instruction-following performance is already unstable across different modalities. 
Beyond model scale, we further observe that explicit CoT mechanisms may interact negatively with instruction adherence, likely due to verbosity and drifting constraints. For example, compared to Qwen3-VL-8B-Instruct, Qwen3-VL-8B-Thinking (marked with \ding{45} in Table~\ref{table_instruction_following}) exhibits a slight but consistent degradation in instruction-following accuracy. A similar trend is observed in Gemini-2.5-Flash, whose instruction-following performance decreases when slow thinking is enabled. 
These findings suggest that instruction-following robustness in sign language understanding is influenced by modalities, architectural and training choices, and may be further affected by the introduction of explicit reasoning.

\begin{figure*}[t]
\centering
\includegraphics[width=1.0 \linewidth]{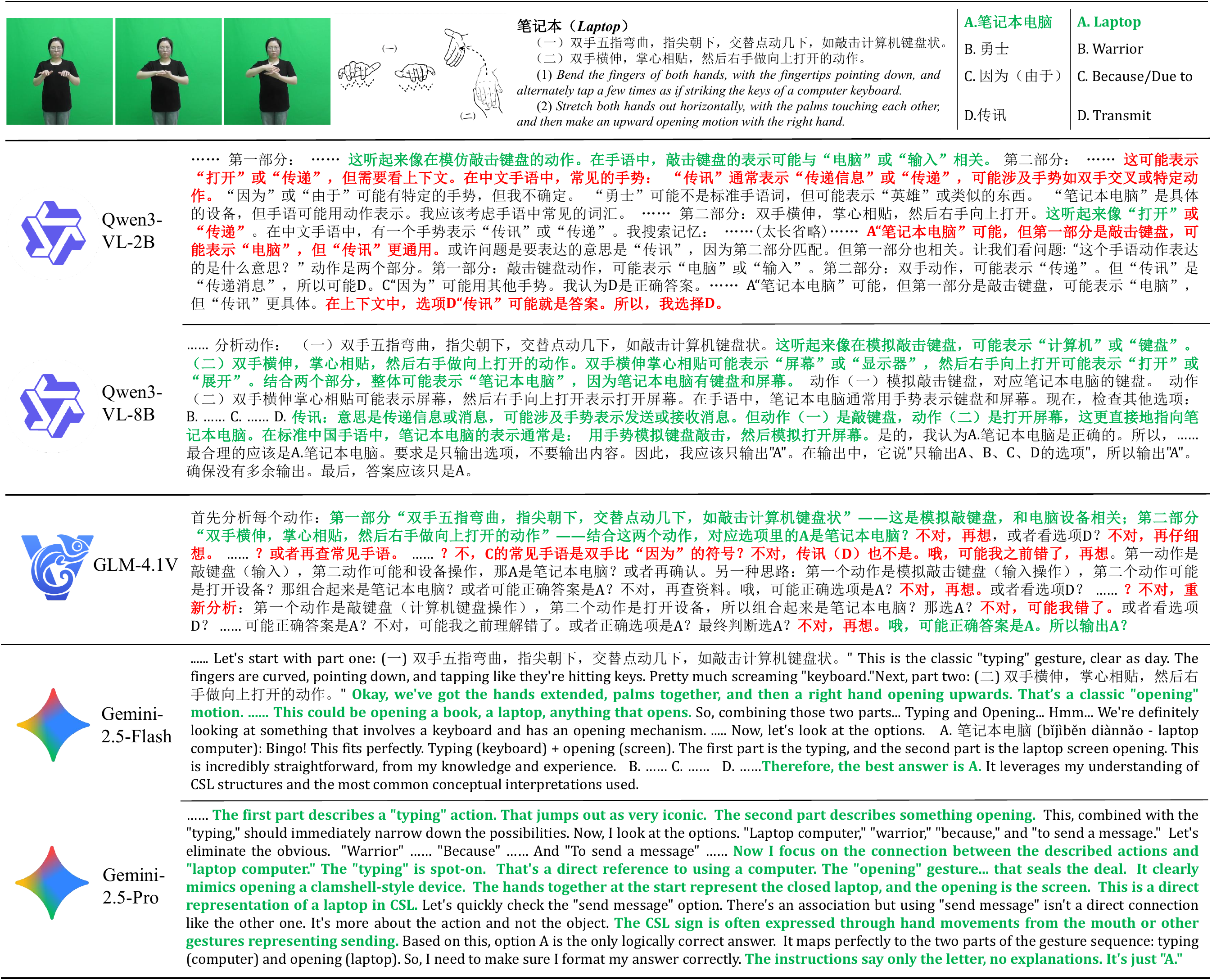}
\caption{Text-based sign language understanding for the lexical concept ``laptop''. Under textual descriptions, models successfully associate the described typing action with the intended concept, demonstrating robust performance when sign language information is abstracted into natural language.}
\label{fig_case_study_1}
\end{figure*}

\subsection{Case Studies}\label{sec_app_case_study}

\begin{figure*}[ht]
\centering
\includegraphics[width=1.0 \linewidth]{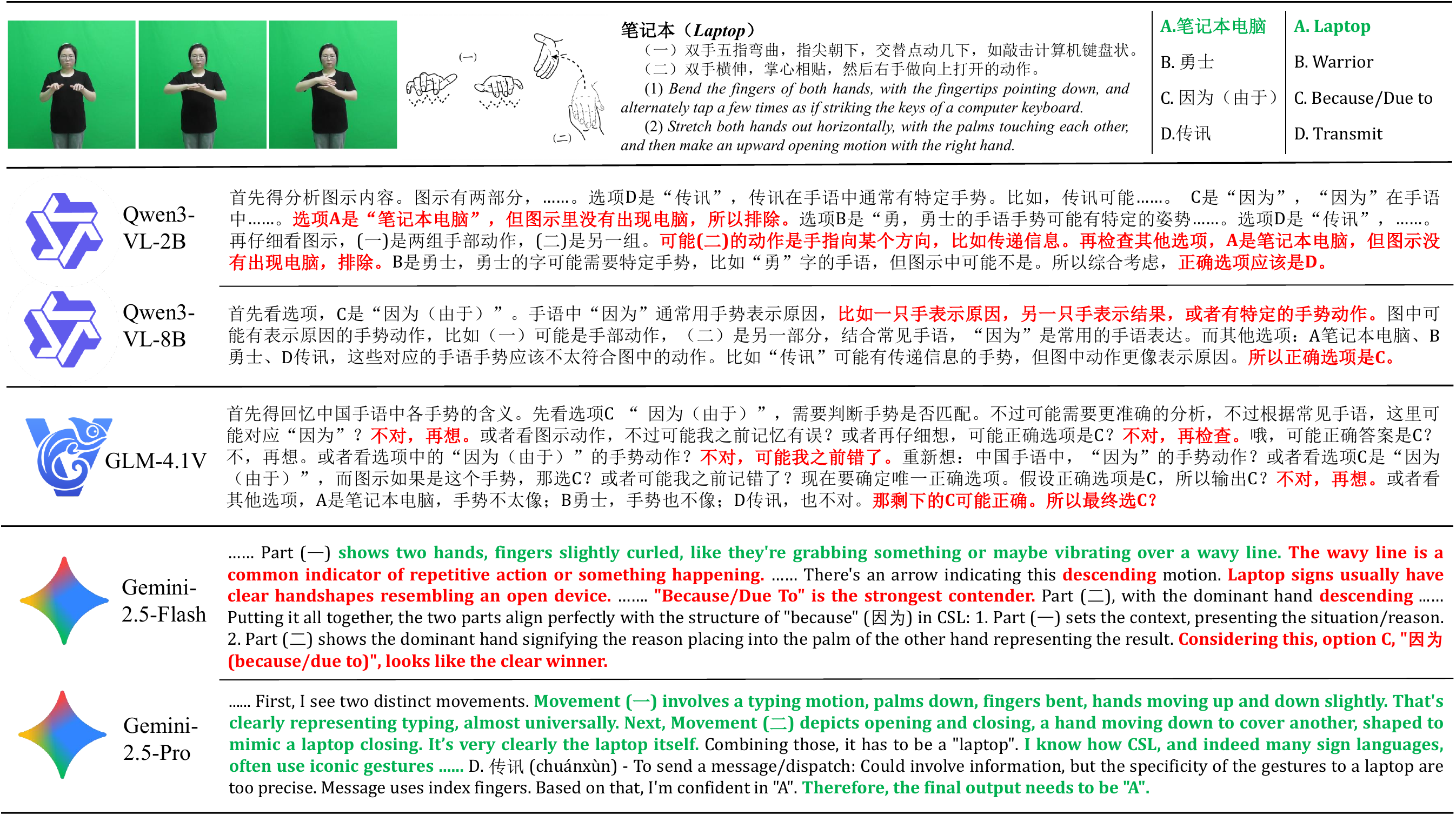}
\caption{Image-based sign language understanding for ``laptop''. With illustrative images, some models correctly infer the target meaning by jointly recognizing multiple visual cues (e.g., ``type'' and ``open'' gestures), highlighting the role of complementary spatial information in sign language interpretation.}
\label{fig_case_study_2}
\end{figure*}
\begin{figure*}[t]
\centering
\includegraphics[width=1.0 \linewidth]{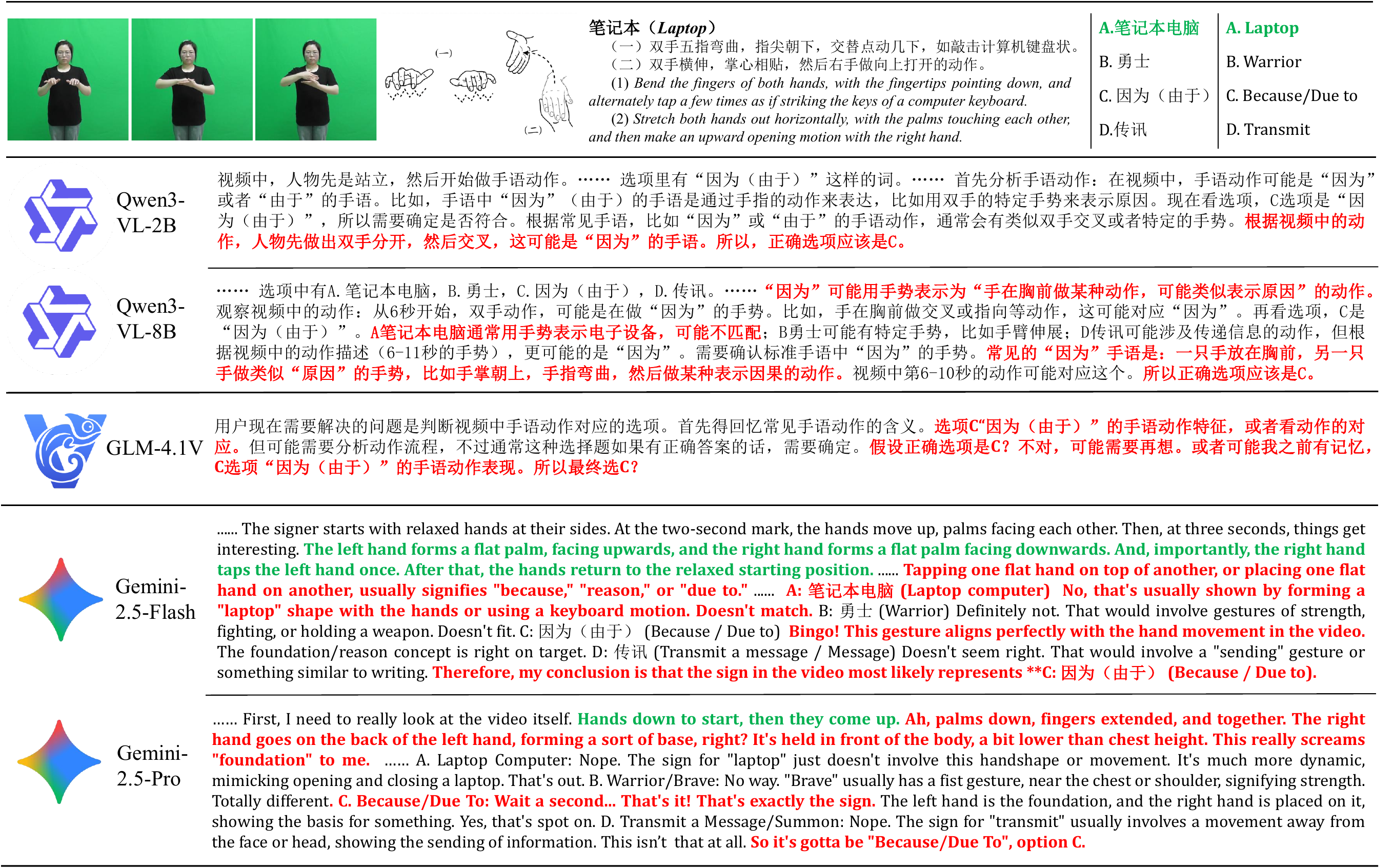}
\caption{Video-based sign language understanding failure for ``laptop''. Despite correctly recognizing the typing motion in a continuous sign language video, models fail to recover the intended meaning, indicating that temporal visual recognition alone is insufficient for reliable sign language understanding.}
\label{fig_case_study_3}
\end{figure*}
\begin{figure*}[t]
\centering
\includegraphics[width=1.0 \linewidth]{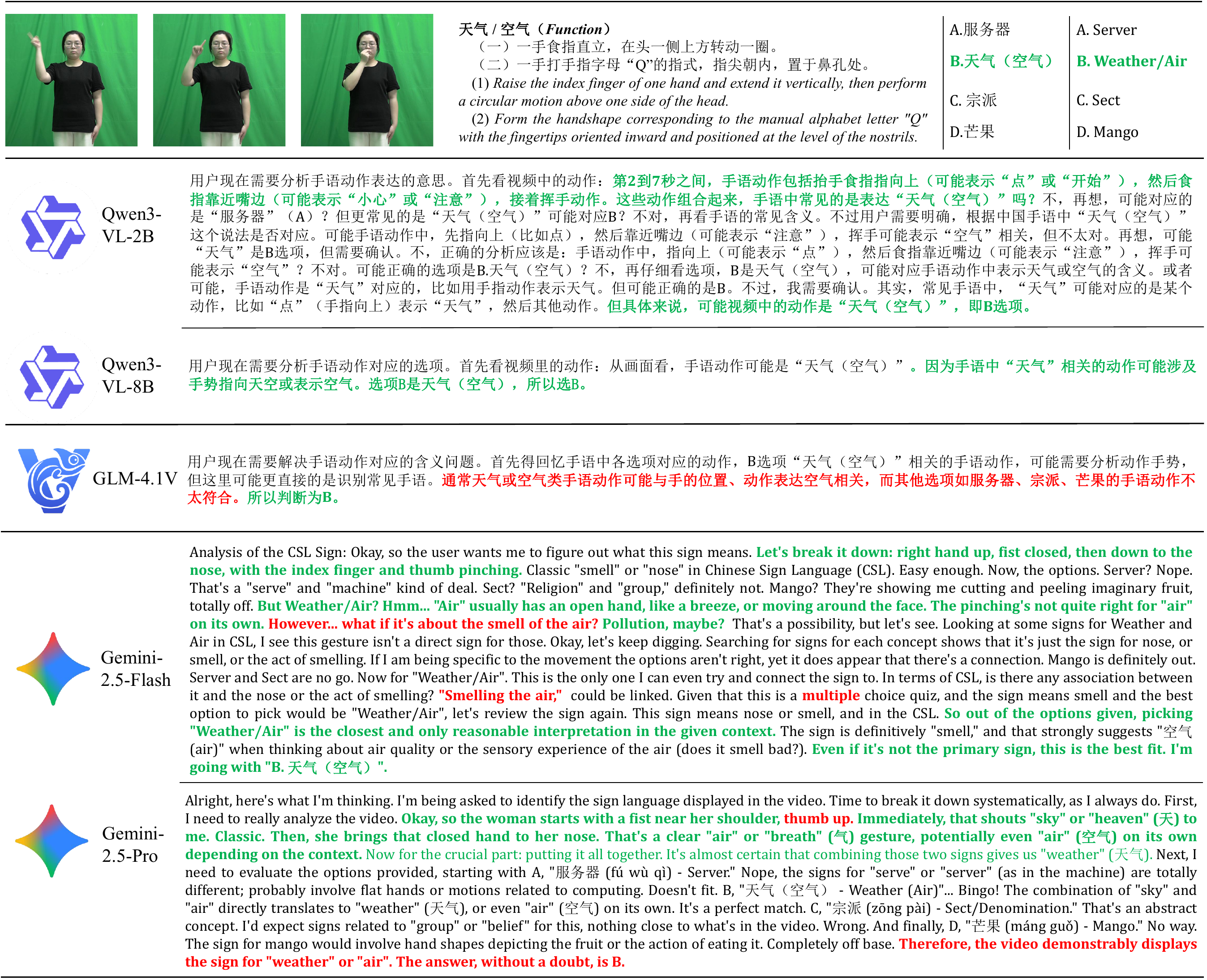}
\caption{Partial semantic recognition without correct contextual integration. Models correctly identify individual sign components (e.g., a nose-related gesture), but fail to integrate them into the intended semantic concept, reflecting the implicit and convention-driven nature of sign language semantics.}
\label{fig_case_study_4}
\end{figure*}
\begin{figure*}[t]
\centering
\includegraphics[width=1.0 \linewidth]{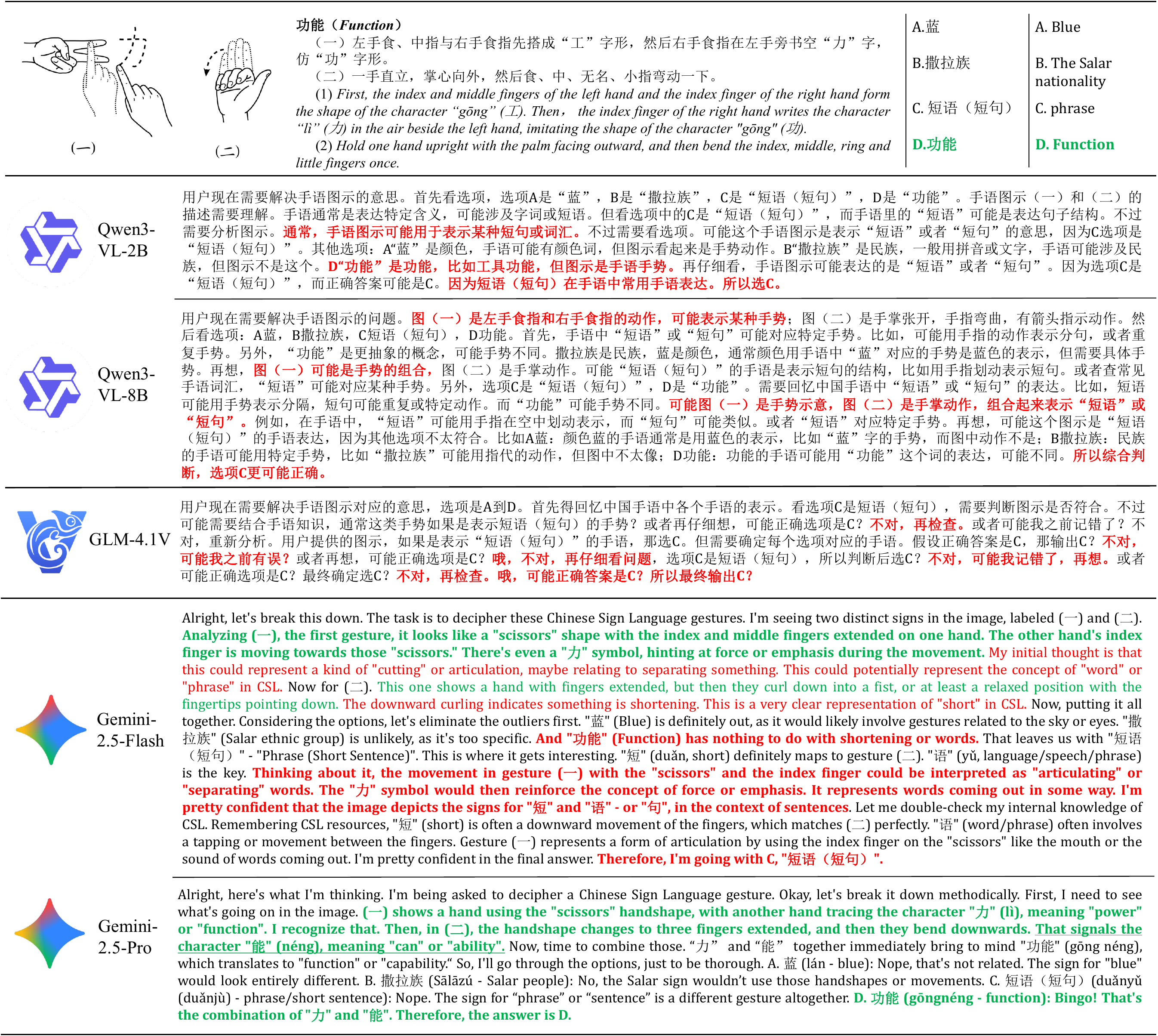}
\caption{Emerging sign-specific symbolic understanding in advanced MLLMs. A stronger model successfully maps a visual sign to the Chinese character ``neng'' (meaning ``can'' or ``ability''), suggesting partial acquisition of sign language knowledge, while still falling short of comprehensive understanding.}
\label{fig_case_study_5}
\end{figure*}

The qualitative cases in Figure~\ref{fig_case_study_1} to Figure~\ref{fig_case_study_5} provide intuitive insights into the sources of observed performance differences and further illustrate the challenges revealed by the CNSL-benchmark.

A primary observation concerns modality sensitivity. For the lexical concept \emph{laptop} (from Figure~\ref{fig_case_study_1} to Figure~\ref{fig_case_study_3}), models consistently succeed under textual descriptions but frequently fail under sign language video inputs, despite correctly recognizing the underlying typing motion. In contrast, image-based inputs sometimes enable correct predictions by exposing complementary cues (e.g., typing combined with an “open” gesture). This contrast indicates that sign language understanding is strongly affected not only by the presence of visual information, but also by how spatial and temporal cues are presented across modalities.
Beyond modality effects, the cases suggest that sign language comprehension often requires integrating perceptual recognition with implicit semantic associations. As shown in Figure~\ref{fig_case_study_4}, Gemini-2.5-Flash may correctly identify individual sign components (e.g., a nose-related gesture corresponding to ``smell'') while failing to connect them to the intended meaning in context. Such failures reflect the fact that sign language is not a simple composition of isolated gestures, but relies on culturally and linguistically grounded conventions that are not always explicitly observable.
At the same time, we observe emerging sign-specific capabilities in stronger models. In Figure~\ref{fig_case_study_5}, Gemini-2.5-Pro correctly maps a visual sign to the Chinese character ``neng'' (meaning ``can'' or ``ability''), indicating partial acquisition of sign knowledge. 

Conclusively, these examples emphasize that the difficulties faced by current MLLMs stem primarily from the intrinsic linguistic and multimodal complexity of sign language, rather than from isolated perceptual errors alone.

\end{document}